\documentclass[10pt,twocolumn,letterpaper]{article}

\usepackage[pagenumbers]{cvpr} 

\usepackage[accsupp]{axessibility} 

\usepackage[dvipsnames]{xcolor}

\definecolor{cvprblue}{rgb}{0.21,0.49,0.74}
\usepackage[pagebackref,breaklinks,colorlinks,citecolor=cvprblue]{hyperref}

\title{CAM Back Again: Large Kernel CNNs from a Weakly Supervised Object Localization Perspective}

\author{Shunsuke Yasuki, Masato Taki\\
Graduate School of Artificial Intelligence and Science, Rikkyo University, Japan\\
{\tt\small Project Page: \href{https://github.com/snskysk/CAM-Back-Again}{https://github.com/snskysk/CAM-Back-Again}}
}

\begin{document}
\maketitle
\begin{abstract}
Recently, convolutional neural networks (CNNs) with large size kernels have attracted much attention in the computer vision field, following the success of the Vision Transformers. Large kernel CNNs have been reported to perform well in downstream vision tasks as well as in classification performance. The reason for the high performance of large kernel CNNs in downstream tasks has been attributed to the large effective receptive field (ERF) produced by large size kernels, but this view has not been fully tested. We therefore revisit the performance of large kernel CNNs in downstream task, focusing on the weakly supervised object localization (WSOL) task. WSOL, a difficult downstream task that is not fully supervised, provides a new angle to explore the capabilities of the large kernel CNNs. Our study compares the modern large kernel CNNs ConvNeXt, RepLKNet, and SLaK to test the validity of the naive expectation that ERF size is important for improving downstream task performance. Our analysis of the factors contributing to high performance provides a different perspective, in which the main factor is feature map improvement. Furthermore, we find that modern CNNs are robust to the CAM problems of local regions of objects being activated, which has long been discussed in WSOL. CAM is the most classic WSOL method, but because of the above-mentioned problems, it is often used as a baseline method for comparison. However, experiments on the CUB-200-2011 dataset show that simply combining a large kernel CNN, CAM, and simple data augmentation methods can achieve performance (90.99\% MaxBoxAcc) comparable to the latest WSOL method, which is CNN-based and requires special training or complex post-processing. 
\end{abstract}    
\section{Introduction}
\label{sec:intro}
Convolutional Neural Networks (CNNs) dominated the computer vision in the 2010s \cite{cnn_lecun,alex_net}. In recent years, however, CNN's position has been threatened by the superior performance of Vision Transformers (ViTs) \cite{ViT_origin,swin_transformer}, which has inspired research efforts to narrow the performance gap between CNNs and ViTs \cite{replknet,slak}. The study of large kernel CNNs is one such example. Multi-head self-attention (MHSA), which is considered one of the key mechanism of Transformers, collects information from a large area on the image. 
It is expected that introducing a larger kernel to CNNs will approximate MHSA's ability to model long-range dependencies.
Large kernel CNNs such as RepLKNet \cite{replknet} have recently attracted a lot of attention because they actually show good performance not only in image classification but also in downstream tasks such as object detection and segmentation.
The success of large kernel CNNs is said to be due to their large size kernel and the large effective receptive field (ERF) \cite{ERF} they produce \cite{replknet,slak}. 
A large ERF can be interpreted as a reflection that the output is able to collect information from a larger area on the input, and is considered a sign that the CNN has acquired the ability to model long-range dependencies similar to MHSA \cite{intern_image}.
The larger ERFs are also expected to improve performance in downstream vision tasks \cite{DeepLab,Rethinking_SS,encodr_decoder_SIS}.
The view that the high performance of large kernel CNNs is due to large ERFs is a compelling argument given the interpretation of ERF size. 

However, there is insufficient validation to support this view, and the following perspectives have not been fully investigated.
First, experimental results on the impact of ERFs on downstream tasks are insufficient; it is not certain whether larger ERFs improve downstream task performance. Second, there is a lack of research on factors other than ERFs. For example, ViTs with large ERFs do not always perform well in downstream tasks due to the breakdown of spatial coupling of images caused by patch embedding and the lack of class identifiability in the attention map \cite{TS_CAM}.
Given these facts, it remains difficult to position the large ERF as a main factor based solely on the results of existing studies, so it is worth further investigating from a new angle what factors are responsible for the performance of large kernel CNNs.

To explore the role of large kernel CNNs from a new perspective,
we focus on the weakly supervised object localization (WSOL) task as a downstream task. WSOL is a task that aims to perform localization using only image-level class labels and is of interest because of its potential to reduce annotation costs.
This task is not a fully supervised task; moreover, it requires discriminative localization of object classes, which requires the generation of high-quality feature maps. 
In addition, CNN-based WSOLs often suffer from partial localization problems \cite{TS_CAM}, making feature maps a challenge to improve. 
We consider that testing large kernel CNNs with WSOL as the downstream task, which is subject to these difficulties, will allow for investigations that are sensitive to the nature of the feature maps.

Based on these ideas, this paper examines the validity of the existing view that ERF size is important for improving the performance of downstream tasks by comparing latest CNNs. In addition, we analyze other high performance factors behind it. Our main contributions are as follows:

\begin{quote}
 \begin{itemize}
  \item We present a new perspective by testing the common but little tested conjecture that ERFs are the main factor of the high performance of large kernel CNNs. Specifically, we show that it is difficult to position the size of ERFs as the main factor of performance improvement, at least for WSOL, and that the high performance of large kernel CNNs is due to improved feature maps.
  \item We find that the feature map improvements described above are effective in avoiding the CAM problem of activating local areas of objects, which has long been discussed in WSOL.
  \item WSOL experiments on the CUB-200-2011 dataset show that simply combining a latest large kernel CNN with the most classic WSOL method (CAM) and simple data augmentation techniques can achieve performance comparable to state-of-the-art CNN-based WSOL methods. 
  Furthermore, the new simple proposed method based on our findings outperforms the CNN-based state-of-the-art score by 0.43\%.
 \end{itemize}
\end{quote}

\section{Related Work}
\label{sec:related}

\subsection{Deep Convolutional Neural Networks}

Research on large convolutional kernels \cite{GCN,involution} has left behind problems such as poor classification performance and saturation of benefits due to increased kernel size. In recent years, however, inspired by the success of Vision Transformers, there has been a growing movement to re-examine CNN architectures. 
ConvNeXt \cite{convnext} introduced 7x7 convolution kernel that is larger than the previous CNN,
influenced by Swin Transformer's 7-12 size window strategy \cite{swin_transformer}.
This allowed the CNN to outperform the Swin Transformer. RepLKNet \cite{replknet} achieved large kernel models up to 31×31 by training large and small kernels in parallel, and SLaK \cite{slak} achieved large kernel models over 51×51 by using a training method with vertical and horizontal rectangular kernels. These can be seen as studies that improve CNNs into models with long-range dependencies, such as Vision Transformers, in terms of kernel size expansion. Hornet \cite{hornet} is an attempt to give CNNs long-range dependencies through gated convolution and recursive design. We study the essential factors that enable large kernel CNNs, which have been increasingly successful in recent years, to perform well in downstream tasks. The view that ERF is the main factor in these high performances is commonly held, but the validity of this view has not been adequately verified. For example, 
 \cite{demystify_vitcnn} investigate the relationship between ERFs and downstream tasks (object detection) and argue that larger ERFs do not necessarily improve the performance of downstream tasks and that the benefits of larger ERFs may saturate as the model scales up. While these views are encouraging and support our argument, the comparison aspect of the Spatial Token Mixer (STM) is strong and differs from the direction of our research.

\subsection{Weakly Supervised Object Localization (WSOL)}

Weakly supervised object localization (WSOL) is a task that aims to localize objects from only image-level labels, and localization is usually performed using a classifier. A representative study of WSOL is CAM \cite{CAM}, which generates heat maps for localization from a model with global average pooling. However, CAMs generated from CNNs trained for classification tend to localize and activate small discriminative portions of objects, and various improvements to this problem have been investigated. Improvements to solve some problems in CAM include data augmentation strategies \cite{Hide-and-seek,ACoL,CutMix,ADL,Adversarial_Erasing}, devising training methods \cite{SPG,I2C,TS_CAM,LCTR,SCM}, specialized post-processing methods using only trained classifiers \cite{grad_cam,grad_cam_pp,layer_cam}, and higher resolution heatmaps \cite{DRN,high_reso_cam,F_cam}, Classification and localization separation \cite{PSOL,SPOL,DiPS}, and other approaches. Following this stream of improvements, we re-examine CAM, a classic and problematic method, in the light of modern CNNs.

\section{Performance Testing of Large Kernel Models in WSOL Tasks}
\label{sec:wsol_test}

By focusing on WSOL as a downstream task, this study aims to give a new perspective on the essential reasons for the performance of modern CNNs.
To the best of our knowledge, there are no research reports on WSOL tasks using large kernel CNNs, so testing the performance of large kernel CNNs in WSOL tasks is also a novel effort.

\begin{figure}[t]
  \centering

    \includegraphics[width=1\linewidth]{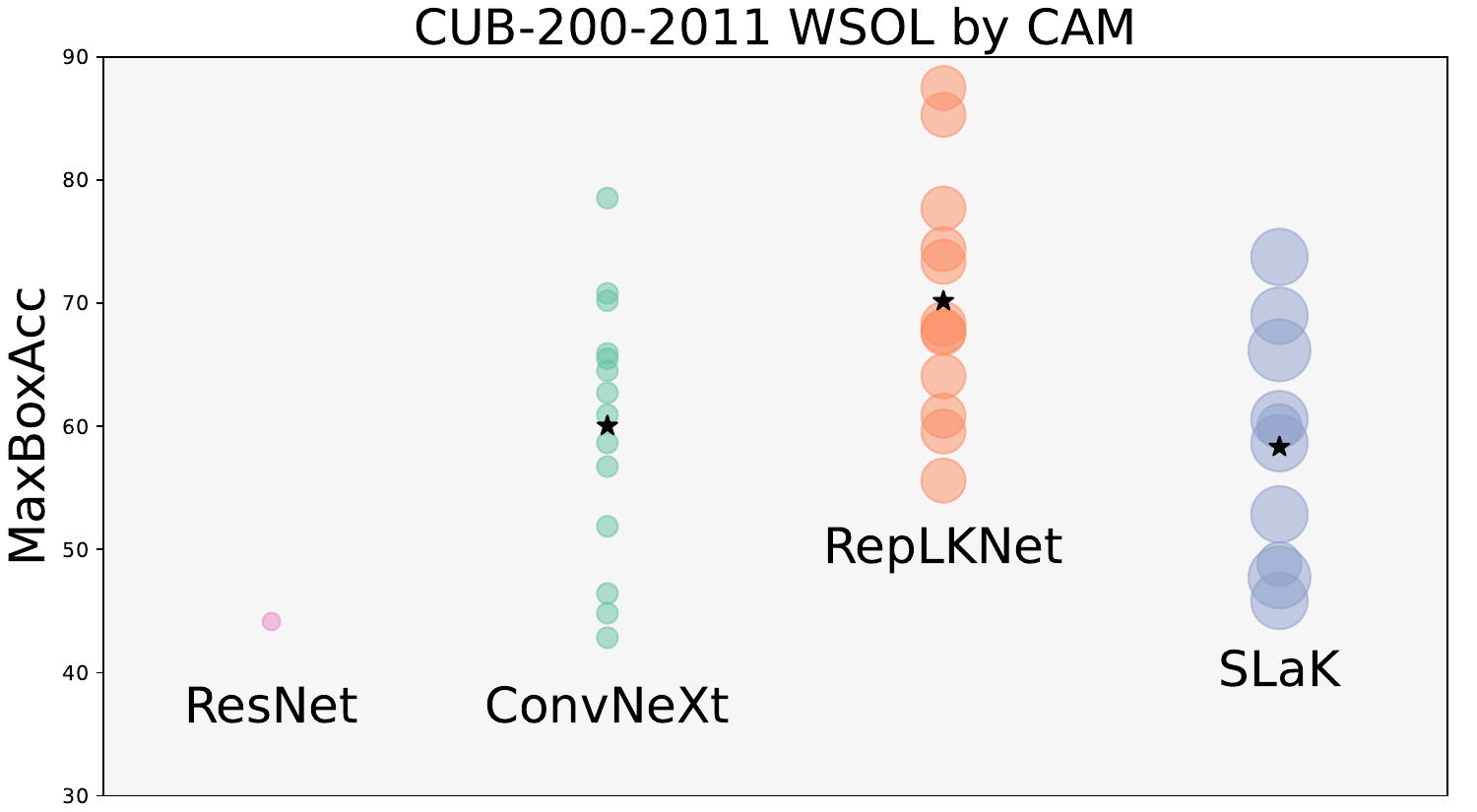}
    \vspace{-1.5em}
    
   \caption{
   WSOL scores (MaxBoxAcc) for CNNs on the CUB-200-2011 dataset. Marker size indicates kernel size. All models pretrained on ImageNet1K are fine-tuned for classifying the CUB-200-2011 dataset. For ConvNeXt, RepLKNet, and SLaK, scores are measured for various models with different pre-training data, pixel size for fine tuning, and number of model parameters.
    }
   \label{fig:fig001}
    \vspace{-1em}
\end{figure}

\cref{fig:fig001} shows the WSOL scores for the traditional CNN (ResNet50 \cite{resnet}) and the latest CNNs on the CUB-200-2011 dataset. As a WSOL method, we use CAM, which is often adopted as a baseline, and the evaluation method is MaxBoxAcc \cite{eval_wsol_right}. All of these models, trained 400 epochs, are not WSOL-optimized (classification-optimized) but show significant overall improvements simply by using the latest CNN as the backbone model. See \cref{sec:app_train_config,sec:app_wsol_eval} for details on fine-tuning and WSOL evaluation.

The RepLKNet model with the highest score in \cref{fig:fig001} is the 31B1K384 in \cite{replknet}, fine-tuned using the same data augmentation method as in the pre-training. The fine-tuning was done with 100 epochs instead of the 400 epochs normally used in RepLKNet and recorded a MaxBoxAcc of 90.99.
\cref{tab:table001} shows our best score and scores of other CNN-based WSOL methods. Our CAM with a RepLKNet backbone scores significantly better than WSOL-optimized CAMs with a ResNet backbone, beating all pre-2021 methods. 
Our model also scores competitively with the newer WSOL methods. Because of various problems with CAM, such as the tendency to activate only local regions of objects, various improvements have been attempted, such as those shown in
\cref{tab:table001}. Those improvements range from data augmentation strategies to more complex training methods and post-processing. However, we have shown 
that simply combining latest large kernel CNNs with the most classic WSOL methods (CAM) and simple data augmentation methods can achieve performance comparable to state-of-the-art CNN-based WSOL methods. In the following sections, we examine the reasons for the performance improvement from several perspectives.

\begin{table}[!htp]\centering
\scriptsize
\begin{tabular}{lrrrr}\toprule
\textbf{Model} &\textbf{Backbone} &\textbf{MaxBoxAccV1} &\textbf{V2} \\\midrule
CAM(cvpr,2016) \cite{CAM} &ResNet &73.2 &63 \\
HaS(iccv,2017) \cite{Hide-and-seek} &ResNet &78.1 &64.7 \\
ACoL(cvpr,2018) \cite{ACoL} &ResNet &72.7 &66.4 \\
DDT(cvpr,2021) \cite{DDT} &VGG16 &84.55* &-- \\
F-CAM(+CAM) (wacv,2022) \cite{F_cam} &ResNet &90.3 &79.4 \\
F-CAM(+LayerCAM) (wacv,2022) &ResNet &92.4 &82.7 \\
F-CAM(+LayerCAM) (wacv,2022) &VGG &91 &84.3 \\
CREAM(cvpr,2022) \cite{cream} &InceptionV3 &90.43* &64.9* \\
BridgeGAP(cvpr,2022) \cite{bridge_gap} &VGG &93.17* &80.1* \\
\textbf{Ours} &RepLKNet &90.99 &76.72 \\
\bottomrule
\end{tabular}
    \vspace{-1em}
\caption{
    Summarizes the best score of our CAM with RepLKNet backbone and the scores of other CNN-based WSOL methods for WSOL on the CUB-200-2011 benchmark dataset. * indicates that the threshold selection for bounding box generation has not been calculated using the different dataset (fullsup dataset \cite{eval_wsol_right}) than the training data or the final prediction data.
}\label{tab:table001}
\end{table}

\section{Analysis of Feature Maps and Weights}
\label{sec:analysis_1}

In this section, we analyze what is responsible for the large improvement in WSOL scores in CAMs with modern CNN backbones, in terms of feature maps and weights.
To do so, we first review the problems that CAM faces in the WSOL task.
Next, we show our observations on how the state of the feature maps and weights of the latest CNN model addresses the existing problems of CAM.

\subsection{Preliminary: Problems Faced by CAMs in WSOL Tasks}

\begin{figure}[t]
  \centering

    \includegraphics[width=1\linewidth]{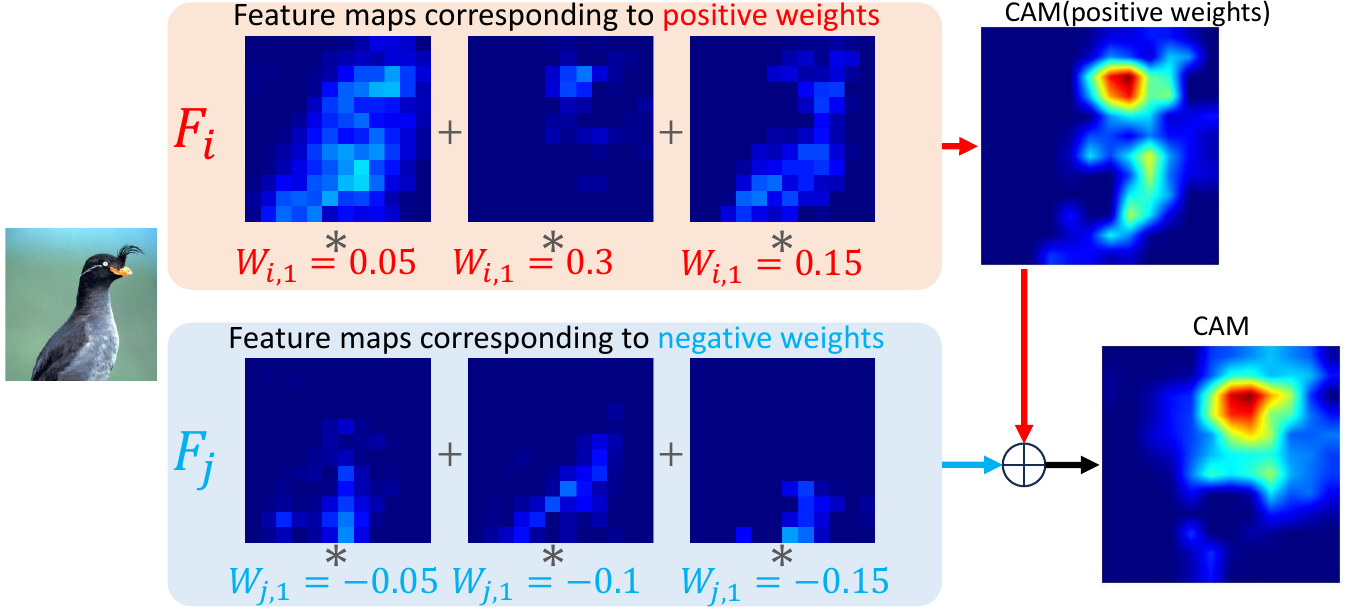}
    \vspace{-1em}
    
   \caption{
   Illustration of the problem that CAMs generated from CNN classifiers tend to locally activate discriminative parts of objects. 
   The figure for $F_i$ represents the problem between the activation area size and the weight size: feature maps with smaller activation regions tend to have larger weights and larger contributions to the CAM. The figure for $F_j$, on the other hand, represents the local activation region problem. This occurs because feature maps corresponding to negative weights activate non-discriminative regions within the object region.
   See \cref{sec:app_cam_pbm} for more information.
    }
   \label{fig:fig002}
    \vspace{-1em}
\end{figure}

Class Activation Map (CAM) \cite{CAM} is a method for localizing object regions in an image that belong to a specific category using CNN with Global Average Pooling (GAP) layer---See \cref{sec:app_cam_calc} for CAM calculations. It is a typical method in WSOL task and is often used as a baseline method. A well-known problem with CAM methods is their tendency to generate locally activated maps. The last CAM in \cref{fig:fig002} looks like a local activation map for a bird's head, which does not allow for correct localization. \cite{rethinking_cam} show how these problems in CAMs occur based on the components of the CAM. Considering their arguments in light of recent WSOL studies, there are two essential problems with CAM. 
First, because the product of activated regions and weights contributes to the logit due to the GAP layer, CNN training tends to increase the weights corresponding to feature maps that activate local regions important for discrimination. This effect enhances the contribution of local feature maps to the CAM.
Second, if feature maps corresponding to negative weights are activated in object areas---especially areas that are not important for classification---, localization is further narrowed down to discriminable areas of the object. \cref{fig:fig002} illustrates these problems using actual feature maps. 

\subsection{Modern CNNs Can Address Existing Problems with CAM}

\begin{figure}[t]
  \centering

    \includegraphics[width=1\linewidth]{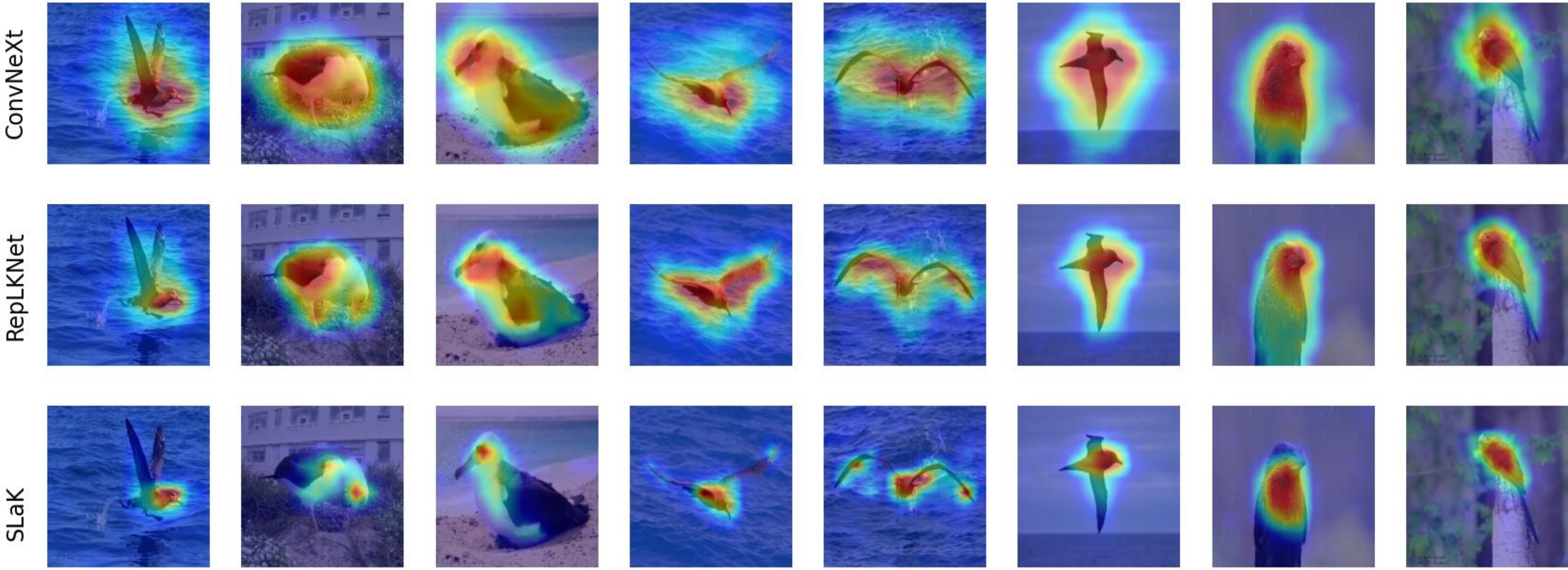}
    \vspace{-1em}
    
   \caption{
   Examples of CAM generation results for the CUB-200-2011 dataset by ConvNeXt, RepLKNet, and SLaK. Among the latest CNNs, ConvNeXt and RepLKNet tend to globally activate the entire object.
    }
   \label{fig:fig003}
    \vspace{-1em}
\end{figure}

While there are two problems that CAM faces in the WSOL task as shown in \cref{fig:fig002}, WSOL scores by CAMs using the latest CNN model have improved dramatically (\cref{fig:fig001}). Furthermore, checking the generated CAMs, we can see that ConvNeXt and RepLKNet, in particular, generate CAMs in which the entire object is activated (\cref{fig:fig003}). These facts suggest that modern CNNs can address the existing problems in CAMs.

\subsection{Solving the Activation Site Problem for Feature Maps Corresponding to Negative Weights}

\begin{figure}[t]
  \centering

    \includegraphics[width=1\linewidth]{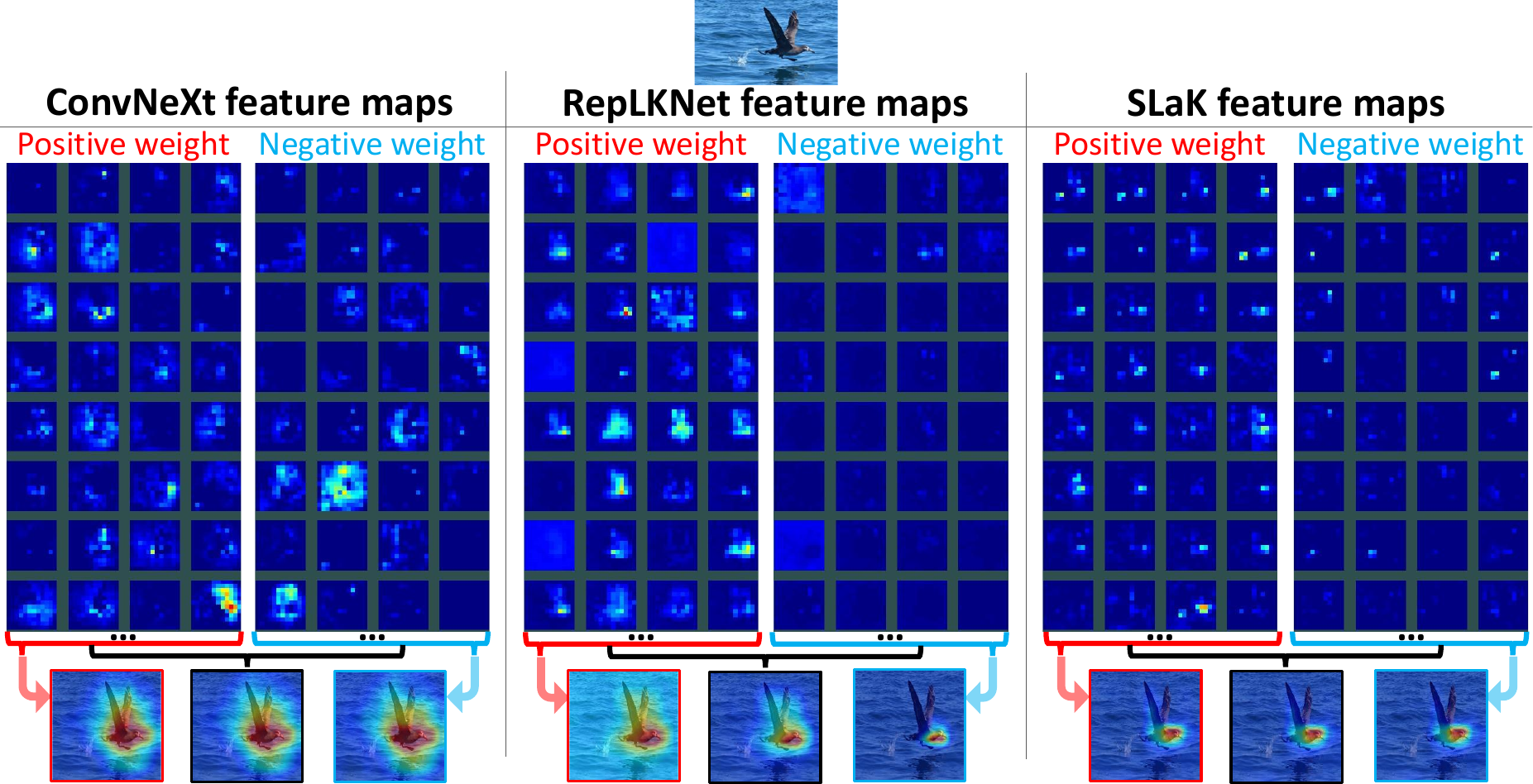}
    \vspace{-1em}    
   \caption{
   Each image group represents $F_{pos}$ and $F_{neg}$ obtained from one input image for each of ConvNeXt, RepLKNet and SLaK. The red-framed heatmap represents the CAM generated from $F_{pos}$ only, the blue-framed heatmap represents the CAM generated from $F_{neg}$ only, and the center heatmap represents the normal CAM.
    }
   \label{fig:fig004}
    \vspace{-1em}
\end{figure}

We show through qualitative and quantitative observations how the latest CNNs address the problem of negative weights shown in $F_j$ in \cref{fig:fig002}. \cref{fig:fig004} show feature maps with large positive and negative weights for three different latest CNNs.
In the following, we refer to the feature maps corresponding to the positive and negative weights as $F_{pos}$ and $F_{neg}$.
ConvNeXt and SLaK have many of $F_{neg}$ activated, while RepLKNet has fewer maps activated. 
$F_{neg}$, which has the effect of suppressing the activation of the final logit for the correct category, should not be activated in the classification task. 
Thus, the RepLKNet feature maps are in a desirable state from the classification task perspective. Such a deactivated state is also desirable for WSOL tasks to avoid this localization problem. RepLKNet seems to avoid this problem by its tendency to automatically deactivate $F_{neg}$.

\begin{table}[!htp]\centering
\scriptsize
\begin{tabular}{lrrrrr}\toprule
&\multicolumn{2}{c}{ImageNet1K MacBoxAcc} &\multicolumn{2}{c}{CUB-200-2011 MacBoxAcc} \\\cmidrule{2-5}
Feature map weights &\textcolor{red}{positive} &\textcolor{blue}{negative} &\textcolor{red}{positive} &\textcolor{blue}{negative} \\\midrule
ConvNeXt &61 &61.2 &67.3 &66.6 \\
RepLKNet &60.1 &7.4 &79 &30.9 \\
SLaK &55.5 &54.5 &66.6 &65.7 \\
\bottomrule
\end{tabular}
    \vspace{-1em}
\caption{
    Measured WSOL scores for CAMs generated from only $F_{pos}$ (positive) and from only $F_{neg}$  (negative). Experiments on two benchmark datasets, ImageNet-1K and CUB-200-2011.
}\label{tab:table002}
    \vspace{-1em}
\end{table}

On the other hand, $F_{neg}$ can still have a negative impact on CAMs in ConvNeXt and SLaK.
We therefore check the CAMs generated from $F_{pos}$ and $F_{neg}$ only. These are illustrated as CAMs in the red and blue boxes at the bottom of \cref{fig:fig004}.
In RepLKNet, we can see that there is a big difference between the $F_{pos}$ and $F_{neg }$ cases. On the other hand, for ConvNeXt and SLaK, there is no such difference.
This phenomenon is also evident in the WSOL scores (\cref{tab:table002}). 
These results suggest that in ConvNeXt and SLaK, the activation patterns of $F_{neg}$ are strongly aligned with those of $F_{pos}$. Thus, contrary to what one would expect from \cref{fig:fig002}, the problems that $F_{pos}$ and $F_{neg}$ have, respectively, no longer occur independently. And if, as in ConvNeXt, the CAM generated from $F_{pos}$ tends to globally activate the entire object, problems on the $F_{neg}$ side are less likely to occur, thus avoiding the CAM problem as a result.

\subsection{Resistance to Activation Area Size and Weight Size Problems}
\label{sec:resistance_}

\begin{figure}[t]
  \centering

    \includegraphics[width=1\linewidth]{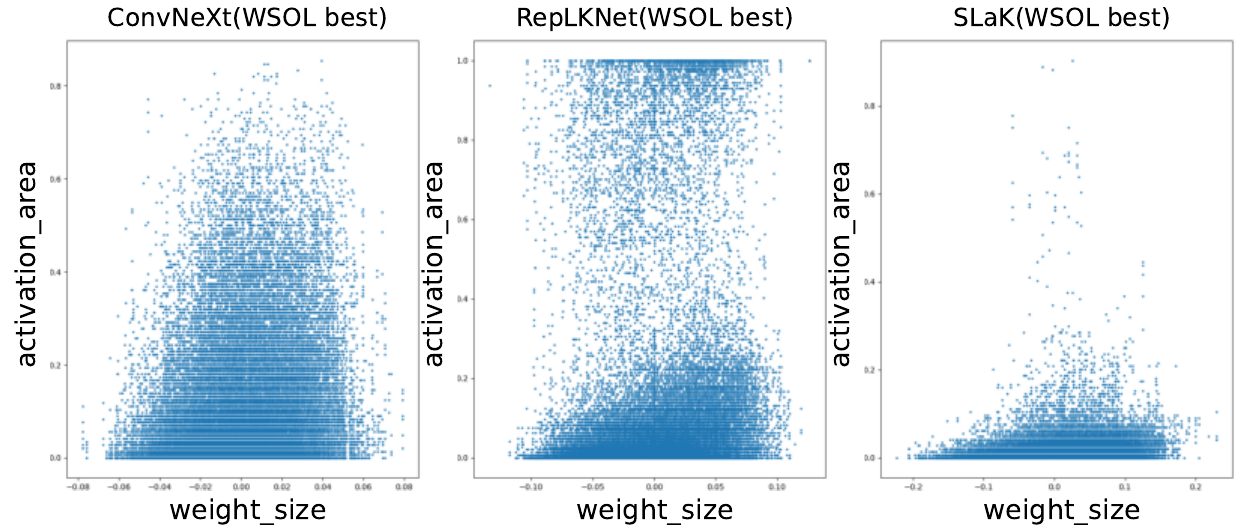}
    
   \caption{
   Relationship between activation areas and weights of feature maps. The activation area is calculated by binarizing the feature map with a threshold of 10 and calculating the percentage of pixels that exceed the threshold. Experiments on ConvNeXt, RepLKNet, and SLaK WSOL top-score models fine-tuned on the CUB-200-2011 dataset. See \cref{sec:app_fmap_ana} for results in the non-best-scoring models.
    }
   \label{fig:fig005}
    \vspace{-0.5em}
\end{figure}

In this section, we present an analysis of how modern CNNs avoid the problem for $F_i$ in \cref{fig:fig002}. First, we check the relationship between the activation areas and weights of the feature maps (\cref{fig:fig005}). Assume that the larger (smaller) the size of the activation region, the smaller (larger) the weights are, regardless of whether they are positive or negative. 
In this case, \cref{fig:fig005} is expected to draw a mountain-like shape. 
The experimental results \cref{fig:fig005} show that the ConvNeXt and SLaK plots have the expected mountainous shape.
ConvNeXt is more peaked than SLaK, so it tends to generate more global CAMs by increasing the activation area of the feature maps.
On the other hand, RepLKNet shows a pillar-like shape rather than the undesirable mountainous shape.

\begin{figure}[t]
  \centering

    \includegraphics[width=1\linewidth]{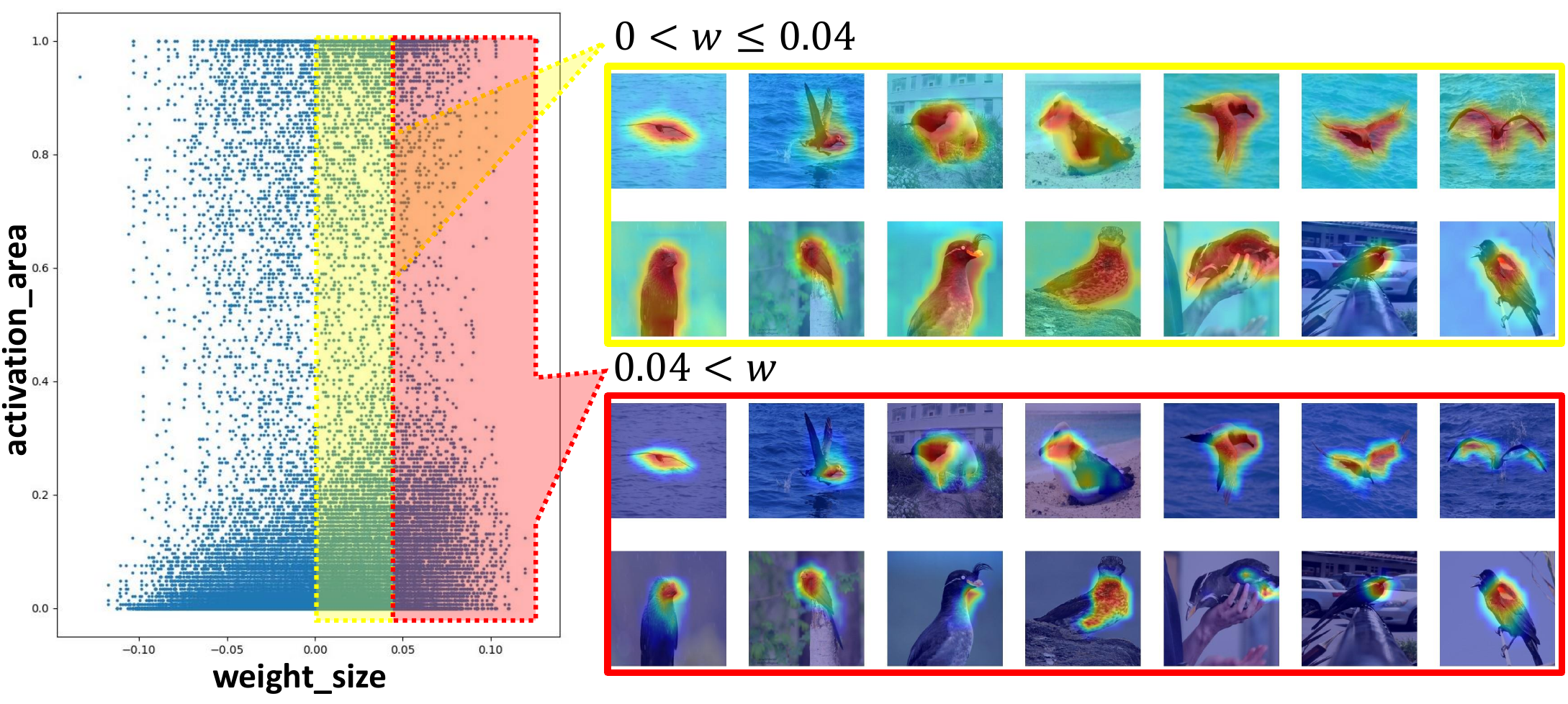}
    
   \caption{
   The illustration compares CAMs generated from feature maps corresponding to small weights---$0<w\leqq0.04$  (yellow box)---and those generated from feature maps corresponding to large weights---$0.04<w$ (red box)---for RepLKNet.
    }
   \label{fig:fig006}
    \vspace{-1em}
\end{figure}

However, as shown in \cref{fig:fig006}, even in RepLKNet, feature maps corresponding to small weights tend to activate the CAM globally, whereas feature maps corresponding to large weights tend to activate the CAM locally.
Thus, the problem of small activation areas giving large weights is not completely solved in RepLKNet.

Despite these unfavorable conditions, CAMs generated by ConvNeXt and RepLKNet tend to be globally activated. This is likely due to the significantly improved global nature of the feature maps themselves, which reduces the impact from the problems described so far.
To confirm this, we perform a principal component analysis of the resulting feature maps. \cref{fig:fig007} visualizes the features of the first principal component (hereafter referred to as PC1) obtained by principal component analysis of the feature maps generated from the bird images.

From \cref{fig:fig007}, we observe that PC1 in ConvNeXt tends to activate global regions that surround the object somewhat extra widely. RepLKNet's PC1, on the other hand, surrounds the object without waste, giving it an ideal shape for a localization map.
SLaK’s PC1 tends to activate locally, and there is also a noticeable tendency to activate regions that are unrelated to the object.

Furthermore, from the distribution of contribution rate of PC1 \cref{fig:fig008}, we can see that PC1 can explain more than half of the feature map in ConvNeXt and RepLKNet.
Thus, by producing globally activated feature maps, ConvNeXt and RepLKNet avoid the local activation problem due to the size of activation regions and weights.

\subsection{Difference in CAM Quality between RepLKNet and ConvNeXt}
\label{sec:diff_in_cam_quality}

The discussion thus far has shown that the long-discussed problem of CAMs tending to be activated locally is solved in RepLKNet and ConvNeXt.
However, RepLKNet tends to generate CAMs that are optimal for localization, whereas ConvNeXt's CAMs tend to be activated somewhat more broadly for the target. 
The reason for this difference can be understood from observations of the feature maps so far in this paper. 

First, as can be seen from the feature map plot \cref{fig:fig004}, RepLKNet's feature maps look cleaner than ConvNeXt's. Checking $F_{pos}$, we can see that many feature maps activate object regions in RepLKNet, while ConvNeXt's feature maps have more diverse and noisy activation patterns. For quantitative material on this observation, see \cref{sec:app_fmap_complexity}. A similar trend can also be seen in the distribution of PC1 contribution rates (\cref{fig:fig008}). ConvNeXt's PC1 contribution rate tends to be much lower than RepLKNet. This is indicative of the diversity of ConvNeXt's feature map, and such noisy activation is likely to blur the CAM contours, leading to overly broad localization.

Similar results were also obtained in an experiment comparing WSOL scores using the binarized PC1 as a localization map. Interestingly, the original WSOL method using PC1 outperformed the state-of-the-art performance of CNN-based WSOL (\cref{tab:table001}) by 0.43\% (\cref{tab:table_ext001}). See \cref{sec:sup_quality_difference} for details.

\begin{table}[!htp]\centering
\scriptsize
\vspace{-1em}
\begin{tabular}{lrrrr}\toprule
\textbf{Model} &\textbf{MaxBoxAcc} &\textbf{MaxBoxAccV2} &\textbf{IoU threshold} \\\midrule
ConvNeXt+PC1 &83.3 &76.5 &82 \\
RepLKNet+PC1 &\textbf{93.6*} &80.4 &34 \\
SLaK+PC1 &30.5 &38.2 &17 \\
\bottomrule
\end{tabular}
\vspace{-1em}
\caption{
     CUB-200-2011 WSOL scores based on PC1.
}\label{tab:table_ext001}
\vspace{-1.5em}
\end{table}

\section{Analysis of the Relationship between Kernel Size, ERF, WSOL score, etc.}
\label{sec:analysis_2}

With respect to why latest CNN models perform well on downstream tasks, it is an interesting study to investigate in detail the previously simple expectation that ERF size is important for performance improvement.
By analyzing the relationship between ERF size and downstream task scores and other factors, this chapter examines the validity of the commonly held view that large ERF sizes are important for high performance.

\begin{figure}[t]
  \centering

    \includegraphics[width=1\linewidth]{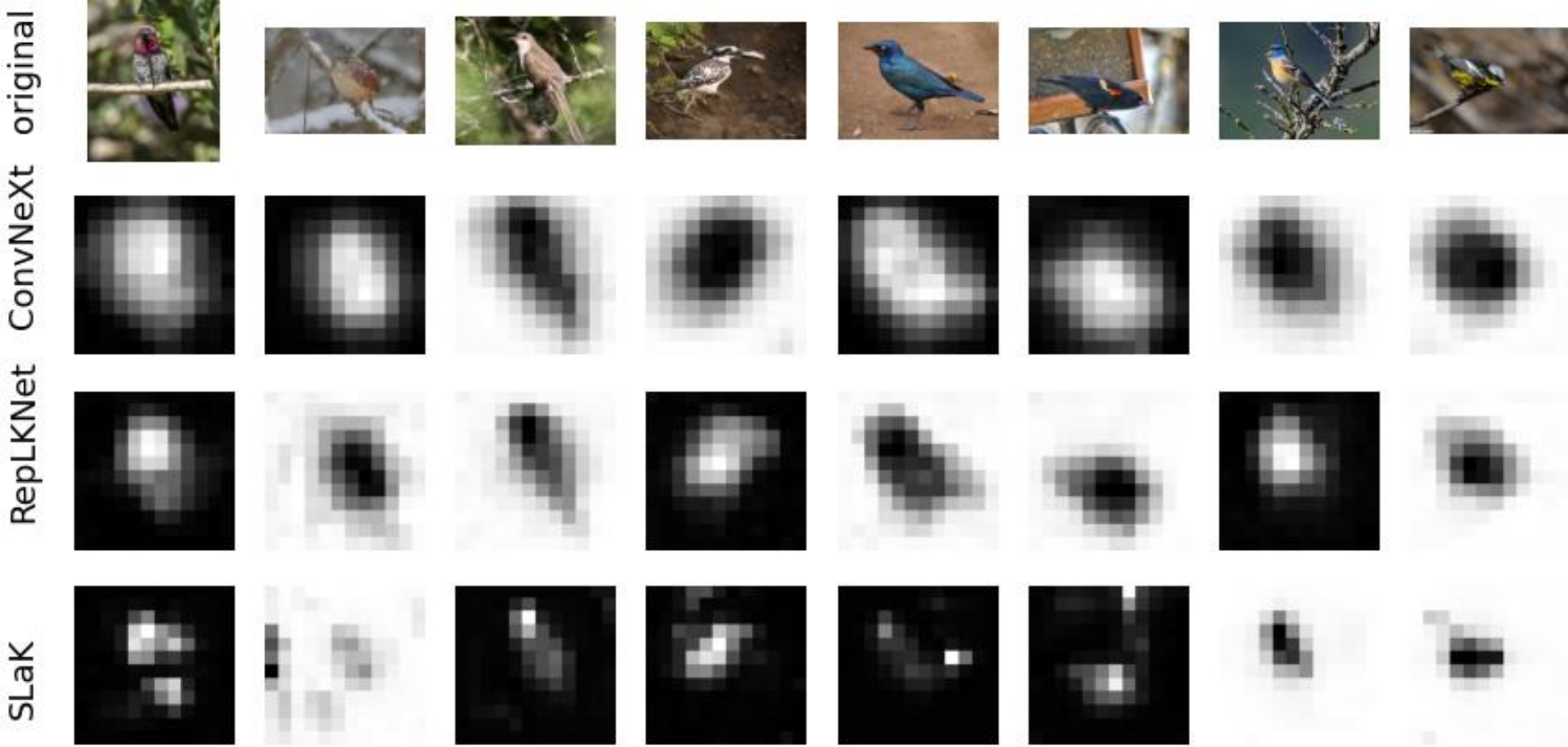}
    \vspace{-1em}    
   \caption{
   PC1 features obtained by principal component analysis of feature maps are visualized. For a subset of the CUB-200-2011 dataset, Feature maps obtained from the final convolution layer of the model with the highest WSOL score are used in the analysis. Pixel size for all models is $12\times 12$.
    }
   \label{fig:fig007}
    \vspace{-1em}    
\end{figure}

\begin{figure}[t]
  \centering

    \includegraphics[width=1\linewidth]{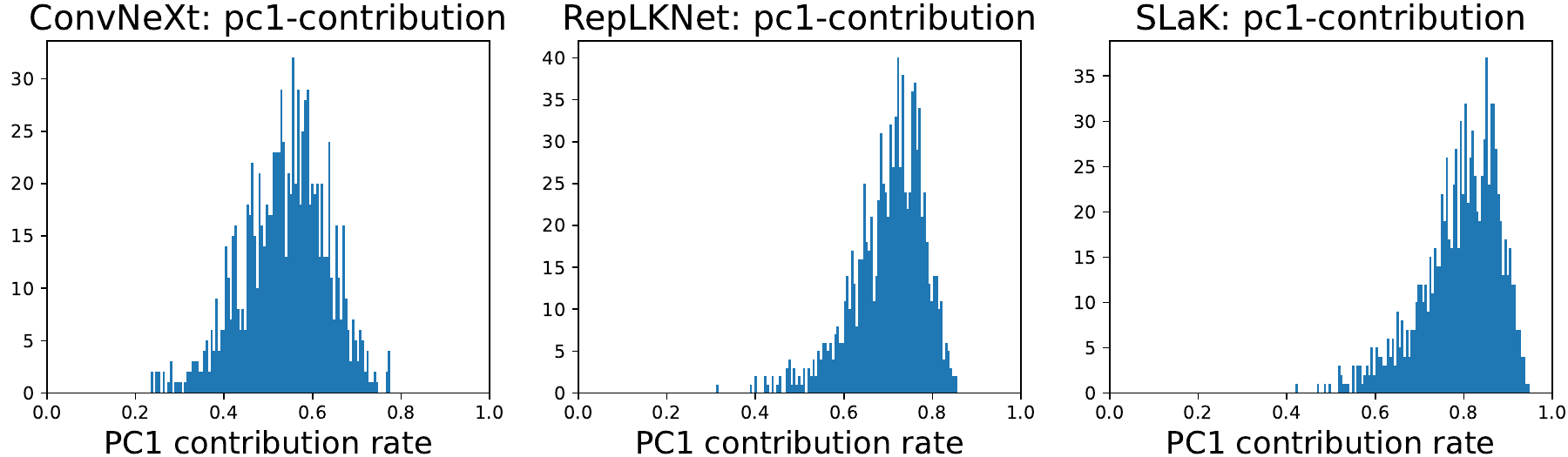}
    \vspace{-1em}        
   \caption{
   The distribution of PC1 contribution rates obtained by principal component analysis of feature maps are visualized. For 1000 fullsup data \cite{eval_wsol_right} of CUB-200-2011, feature maps obtained from the final convolution layer of the model with the highest WSOL score are used in the analysis.
    }
   \label{fig:fig008}
\end{figure}

\begin{figure}[t]
  \centering

    \includegraphics[width=1\linewidth]{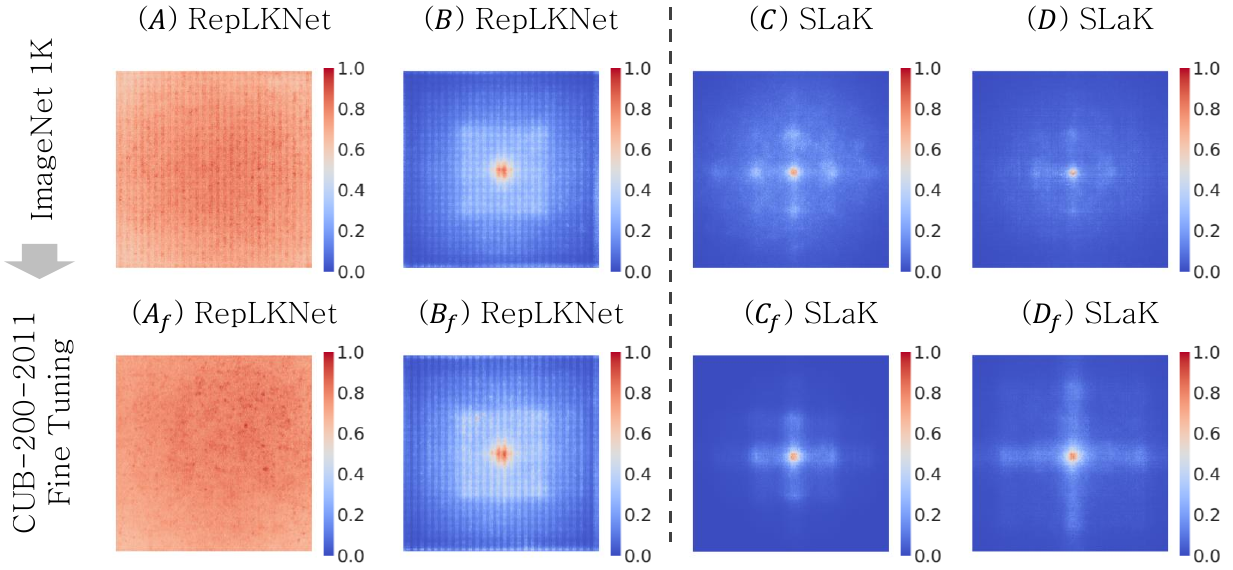}
    \vspace{-1em}        
   \caption{
   $A$, $A_f$, $B$, $B_f$, visualize the effective receptive fields of RepLKNet with a kernel size of 31, $C$, $C_f$, SlaK with a kernel size of 51, $D$, $D_f$, SlaK with a kernel size of 61---A:31B1K224, B:31B22K384, C:SLaKtiny, D:SLaKtiny. 100 CUB-200-2011 images are sampled and resized to $1024\times 1024$ to measure the contribution of pixels on the input image to the center point of the feature map generated in the final convolution layer.
    }
   \label{fig:fig009}
    \vspace{-1em}
\end{figure}

\begin{figure}[t]
  \centering

    \includegraphics[width=1\linewidth]{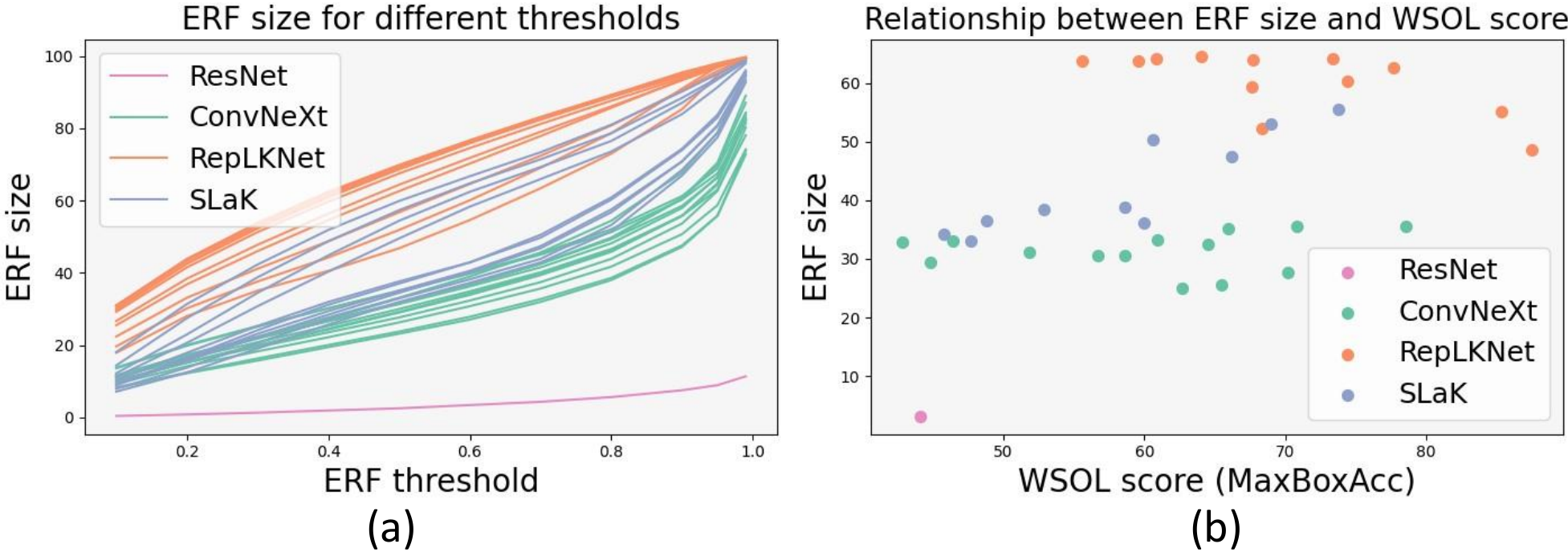}
    
   \caption{
   (a) shows the ERFs of the various fine-tuning models for the CUB-200-2011 dataset. ERF size is measured at 11 different threshold values of [0.1, 0.2, 0.3, 0.4, 0.5, 0.6, 0.7, 0.8, 0.9, 0.95, 0.99]. (b) shows the relationship between WSOL score (MaxBoxAcc) and ERF size. Here, ERF size is the Area Under the Curve (AUC) of (a).
    }
   \label{fig:fig010}
\end{figure}

\subsection{Does the Larger the Kernel Size, the Larger the ERF?}

According to theoretical analysis of ERF \cite{ERF}, ERF grows linearly with kernel size. However, there are many factors that determine the actual size (kernel size, dilated convolution, subsampling, layer depth, skip connections, etc. \cite{ERF}), and a larger kernel size does not necessarily mean a larger ERF. Thus, when comparing RepLKNet with a kernel size of 31 and SLaK with a kernel size of 51, we cannot necessarily say that the latter ERF is larger.
In this regard, \cite{slak} shows that the ERF of the latter is larger.

However, as shown in \cref{fig:fig009}, we found several cases in which the ERF of RepLKNet with a smaller kernel size is larger than that of SLaK. Detailed quantitative verification (\cref{fig:fig010} (a)) also confirms that the ERF size of SLaK is smaller than that of RepLKNet in many cases. The ERF size for the SLaK model group, shown in blue in the graph, is smaller than the ERF size for the RepLKNet model group, shown in orange. Furthermore, the ERF sizes of some ConvNeXt models (green) exceed those of some SLaK models with smaller ERF sizes.
One reason for the relatively small ERF despite the large kernel size of the SLaK model may be due to its architectural design. Reflecting SLaK's use of decomposition into vertical and horizontal rectangular kernels, \cref{fig:fig009} shows that SLaK's ERF shape is more cross-shaped than RepLKNet's ERF.

In summary, the ERFs of the latest CNN models are overall much larger than those of classic CNN models such as ResNet, but we observed that the kernel sizes of these reported models are not necessarily the dominant determinant of ERF size.

\subsection{Does a Larger ERF Result in a Higher WSOL Score?: Analysis of Conditions that Result in a Higher WSOL Score}

\cref{fig:fig010} (b) shows the relationship between ERF size and WSOL score (MaxBoxAcc) for various models. While the overall trend appears to be for larger ERFs to have higher WSOL scores, observations for each group of architectures indicate that the relationship between ERFs and scores is unstable.
For example, RepLKNet is the architecture group with the highest WSOL scores, but within this group, models with higher scores tend to have smaller ERF sizes. Also, some models in ConvNeXt and SLaK have WSOL scores comparable to ResNet, which has a significantly smaller ERF size. These results suggest that the relationship between ERF size and WSOL score is not so simple. See \cref{sec:app_wsol_eval} for MaxBoxAccV2.

\begin{figure}[t]
  \centering

    \includegraphics[width=1\linewidth]{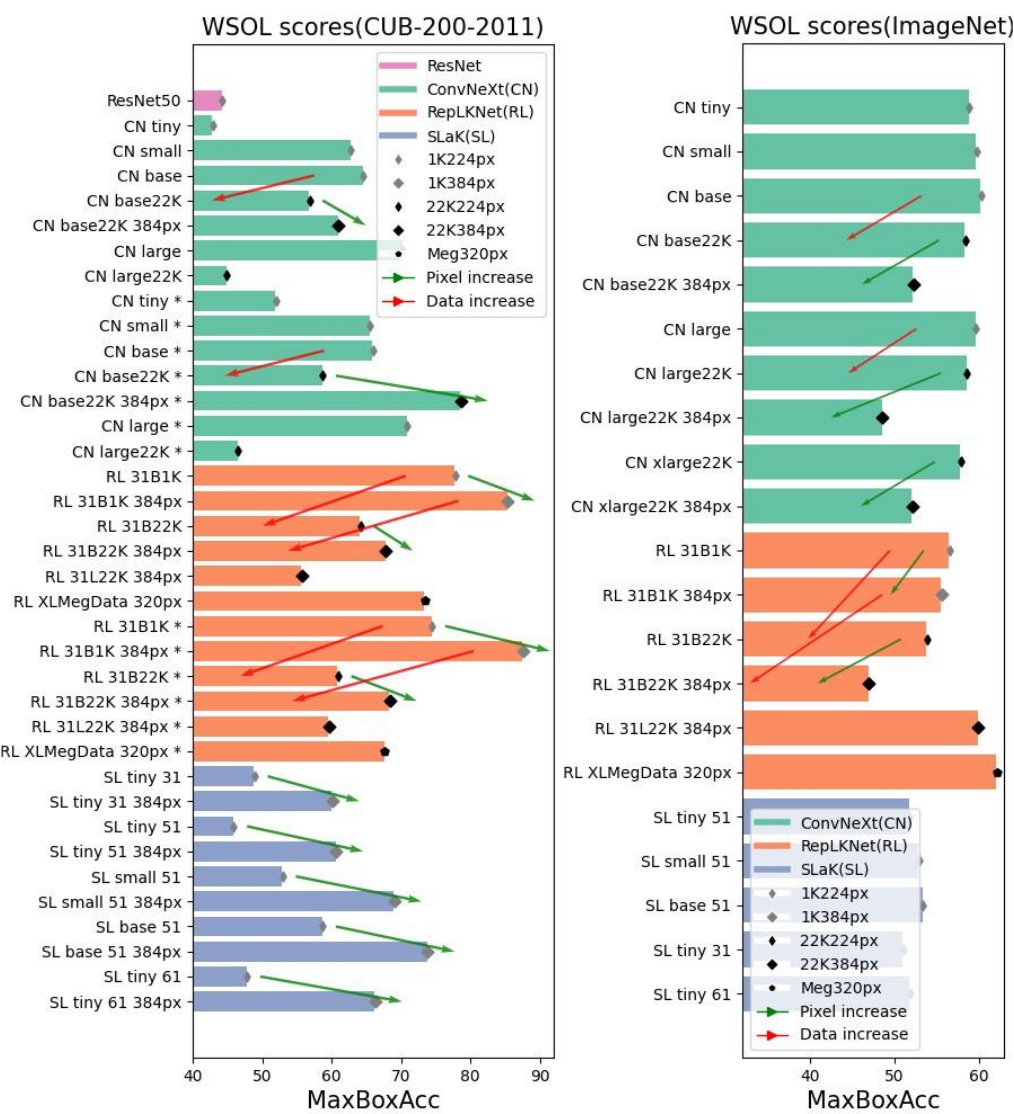}
    
    \vspace{-1em}    
   \caption{
   WSOL scores (MaxBoxAcc) for the CUB-200-2011 dataset on the left and the ImageNet1K dataset on the right. Green arrows connect models that differ only in training pixel size settings, and red arrows connect models that differ only in pre-training settings. * denotes models that introduced the same data augmentation strategy as the pre-training in fine tuning.
    }
   \label{fig:fig011}
\end{figure}

\cref{fig:fig011} also provides insight into the relationship between WSOL scores and training data rather than ERF size.
To summarize what can be seen from the direction of the arrows in each graph in \cref{fig:fig011}, WSOL scores tend to be lower for pre-trained models and for models pre-trained by larger dataset. The WSOL scores for larger training pixel size settings are higher for the CUB-200-2011 dataset and lower for the ImageNet1K dataset.

Thus, the relationship between training data and WSOL scores was seen in some cases with different trends across datasets and in others with consistent trends regardless of the dataset. Other influencing factors, such as changes in training data and training settings, also make the importance of ERFs in performance improvement uncertain. While the overall ERF size increases for ConvNeXt, SLaK, and RepLKNet in that order, ConvNeXt and RepLKNet CAMs are more likely to be activated globally \cref{fig:fig003}. Given these observations, it becomes increasingly difficult to justify the relationship between ERF and performance improvement. Thus, the above results do not support the existing expectation that ERF size is important for the performance of downstream tasks in latest CNN models, and it is difficult to position ERF size as the main driver of performance improvement.
See \cref{sec:app_shape_bias} for additional analysis on ERF.

\section{Discussion}
\label{sec:discussion}

\subsection{Why ConvNeXt and RepLKNet are More Likely to Generate Globally Activated CAMs}

CAM tends to be global for ConvNeXt and RepLKNet and local for SLaK, but the reason for this is not clear. Therefore, in this section we discuss the origin of this trend based on experimental results.

First, what makes the interpretation of CAM trends difficult is the fact that CAM activation is local for SLaK, which has a larger kernel size than RepLKNet, whereas it is global for ConvNeXt, which does not have as large a kernel size. 
For these reasons, it is difficult to expect that the large kernel is the main cause of global CAM. Also, despite the fact that ERF sizes are larger for ConvNeXt, SLaK, and RepLKNet, in that order, ConvNeXt and RepLKNet are the two that achieve large CAM.

\begin{figure}[t]
  \centering

    \includegraphics[width=1\linewidth]{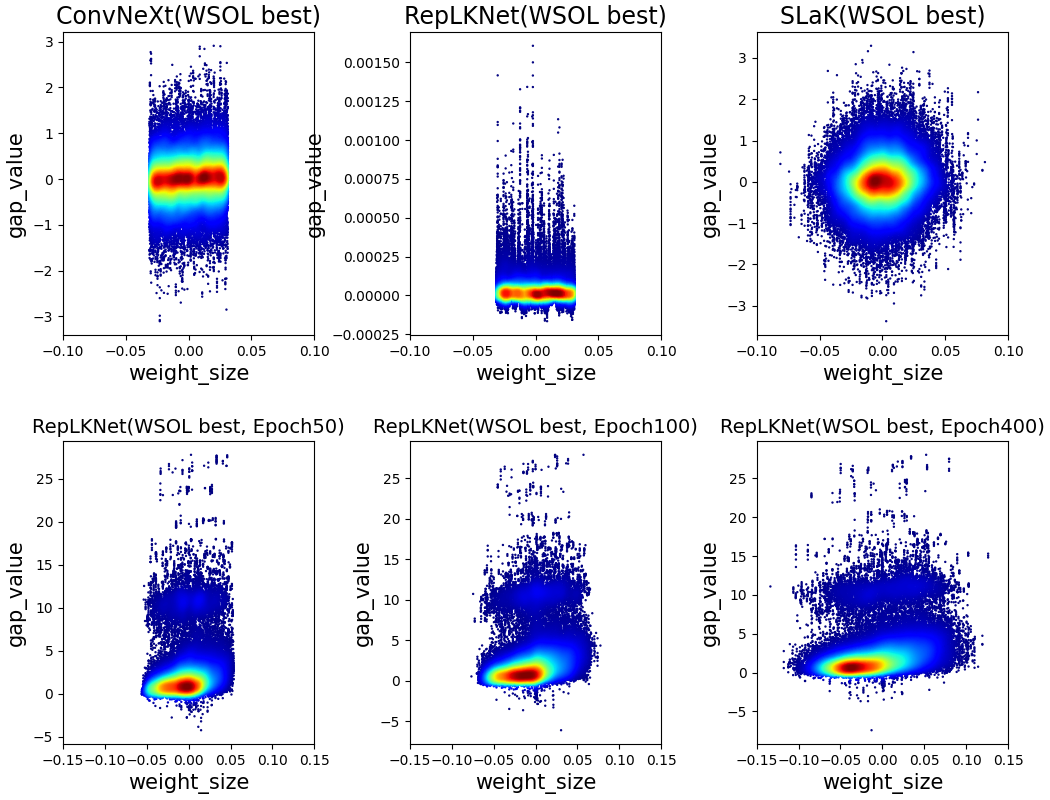}
    \vspace{-1em}    
    
   \caption{ 
   This figure shows the relationship between the GAP values and weights that the feature maps have. The first row shows the results for the three models with the highest WSOL scores in the CUB-200-2011 dataset. The weights are randomly initialized and plotted against 100 feature maps obtained from 100 random images selected from the same dataset.
   The second row represents the 50, 100, and 400 epoch states in RepLKNet of the first row. See \cref{sec:app_fmap_ana} for results in the non-best-scoring models.
    }
   \label{fig:fig012}
\end{figure}

We therefore focus on the inherent properties of each CNN architecture. To see the impact of architectural design, we prepare models with randomly initialized weights and compare and analyze the feature maps obtained from them.
The first line of \cref{fig:fig012} shows the relationship between the GAP values of the feature maps and the weight sizes. This figure is evaluated against feature maps generated from 100 random bird images (CUB-200-2011 dataset) for each of ConvNeXt, RepLKNet, and SLaK with randomly initialized weights.
Each model is selected for achieving the highest WSOL score when fine-tuning on the CUB-200-2011 dataset.

The GAP values reflect the area of the activated region through global pooling.
For SLaK, the relationship between GAP values and weights is distributed as a 2D Gaussian centered at the origin. ConvNeXt and RepLKNet, on the other hand, have a widely distributed bias in the direction of the GAP values. The Gaussian shape of SLaK strongly resembles the general trend of CNNs, where the smaller (larger) the activated regions in the feature map, the larger (smaller) the weights tend to be.
Thus, for the same reasons explained in \cref{sec:resistance_}, SLaK inherently has a tendency to produce locally activated CAMs from the time of initialization.
On the other hand, in the plots for ConvNeXt and RepLKNet, large weights do not necessarily imply small GAP values, thus avoiding the general CNN tendency mentioned above. Thus, already at the time of initialization, it is suggested that ConvNeXt and RepLKNet have a strong tendency to generate globally activated feature maps.

Furthermore, compared to ConvNeXt, RepLKNet's distribution is skewed toward positive GAP values. ConvNeXt's GAP values are centered at 0 and spread evenly in both positive and negative directions. In contrast, RepLKNet's GAP values are most concentrated in positive GAP values and the entire distribution is long in the positive direction.
This characteristic may have resulted in feature maps with even larger GAP values and activation area sizes through training. 
Note that the distribution of RepLKNet changes during the training process, as shown in the second row of \cref{fig:fig012}---50, 100, and 400 epochs from left to right. 
From these plots, we can see that the separation of the RepLKNet distribution into two peaks, upper and lower, observed in \cref{fig:fig005}, occurs during the training process.

\subsection{Impact of Data Augmentation Strategies}

The latest CNN innovations are not limited to those related to architecture. For example, the training of RepLKNet utilizes training techniques such as AdamW optimizer \cite{AdamW}, Stochastic Depth \cite{stochastic_depth}, and data augmentation techniques such as RandAugment \cite{randaugment}, mixup \cite{mixup}, CutMix \cite{CutMix}, Rand Erasing \cite{rand_erasing}. Of these, CutMix is known to be effective in improving the performance of downstream tasks including WSOL. CutMix forces the classifier to focus on a wider range of cues, including non-discriminative parts of the object, through a cut-and-paste operation on the patches. As a result, it is expected to learn a spatially distributed representation \cite{CutMix}. Therefore, the impact of CutMix on WSOL performance of latest CNNs needs to be checked.

A comparison of the differences in WSOL scores after fine-tuning with and without data augmentation processing, including CutMix, showed that the range of variation in WSOL scores with the additional processing ranged from -5.8\% to +4.0\%, with no positive or negative bias (\cref{tab:table003}). 
In addition, the pre-trained model RepLKNet XL MegData 320px is the only ImageNet-1K classification model for which no data augmentation was performed during the pre-training process.
If this model is used as the pre-training model and no data augmentation is applied in fine-tuning, the WSOL score is 73.4\%. This is higher than 7 of the other 11 models that received data augmentation during the pre-training or fine-tuning (or both) process.

Based on these results, we conclude that the architecture itself, more than the data augmentation strategy, contributes to the improved performance of the downstream tasks of the latest CNNs.

\begin{table}[!htp]\centering
\scriptsize
\begin{tabular}{lrrrr}\toprule
Pre-trained RepLKNet model &without DA &with DA &Difference \\\midrule
31B1K 224px &77.68 &74.39 &\textcolor{blue}{-3.30} \\
31B1K 384px &85.30 &87.47 &\textcolor{red}{+2.17} \\
31B22K 224px &64.07 &60.86 &\textcolor{blue}{-3.21} \\
31B22K 384px &67.73 &68.31 &\textcolor{red}{+0.59} \\
31L22K 384px &55.59 &59.58 &\textcolor{red}{+3.99} \\
XLMegData 320px &73.35 &67.59 &\textcolor{blue}{-5.76} \\
\bottomrule
\end{tabular}
\caption{
    Results summarizing the differences in WSOL scores for six RepLKNet models---ImageNet1K classification models---with different training setups, with and without data augmentation (DA) strategies when fine-tuning on the CUB-200-2011 dataset.
}\label{tab:table003}
    \vspace{-1em}    
\end{table}

\section{Conclusion}
\label{sec:conclusion}
In this paper, we found that when combined with latest large kernel CNNs, simple CAM method in WSOL tasks performs well competitively with state-of-the-art CNN-based WSOL methods. In addition, we revisited the main factors behind the high performance exhibited by latest large kernel CNNs from the perspective of the WSOL task. As a result, we present some material that precludes us from placing the size of the ERF as the main factor behind the high performance. We then provide a new view that high performance in WSOL tasks is due to the inherent capability of the architecture  (e.g., structures that facilitate the generation of feature maps with large GAP values or activation regions) and the resulting improvement in feature maps. We also confirmed that the CAM problem of local regions of objects being activated, which has long been discussed in WSOL, is effectively solved by the improved feature map.
Finally, we hope that our research, which evaluates both architecture-specific inherent characteristics and tendencies acquired through training from a WSOL perspective, will lead to a deeper understanding of the properties required for better vision models and facilitate systematic architecture development.

{
    \small
    \bibliographystyle{ieeenat_fullname}
    \bibliography{main}
}

\clearpage
\setcounter{page}{1}
\maketitlesupplementary

\section{Training Configurations}
\label{sec:app_train_config}

In this study, models pre-trained on the ImageNet1K dataset are fine-tuned for the CUB-200-2011 dataset. 
12 fine-tuned models are trained from 6 pre-trained models in RepLKNet, 14 fine-tuned models are trained from 7 pre-trained models in ConvNeXt, and 10 fine-tuned models are trained from 5 pre-trained models in SLaK.
As a preprocessing step for the CUB-200-2011 dataset for training, we resize images to square size of 512 x 512 and then apply center cropping at a ratio of 0.875 to obtain 448 x 448 images.

We first present the details of RepLKNet and ConvNeXt. For RepLKNet fine-tuning, we use the following 6 pre-trained models provided by \cite{replknet}; 31B1K224, 31B1K384, 31B22K224, 31B22K384, 31L22K384, XL. For ConvNeXt fine-tuning, we use the following 7 pre-trained models provided by PyTorch Image Models (timm) library; convnext-tiny, convnext-small, convnext-base, convnext-base-in22ft1k, convnext-base-384-in22ft1k, convnext-large, convnext-large-in22ft1k. With 8 GPUs, NVIDIA Quadro RTX 8000, we train 400 epochs with a batch size of 16 per GPU. The learning rate is $10^{-4}$ 
for the FC layer and $10^{-5}$ 
for the rest. The input resolution is set to the resolution of the pre-trained model, and other settings follow the fine-tuning settings of \cite{replknet}. With or without data augmentation in fine tuning, two fine-tuned models are obtained from one pre-trained model.

Next, we present the details of SLaK. The pre-trained model uses SLaK-T (K=31×31), SLaK-T (K=51×51), SLaK-T (K=61×61), SLaK-S (K=51×51), SLaK-B (K=51×51) provided by \cite{slak} (K is kernel size). They are all trained at 224 x 224 resolution, but the difference in whether the fine-tuning pixel size setting is 224 x 224 or 384 x 384 results in two fine-tuned models from a single pre-trained model. With 8 GPUs, NVIDIA Quadro RTX 8000, we train 400 epochs with a batch size of 16 per GPU. The drop-path is set to 0.1, 0.4, and 0.5 for tiny, small, and base, respectively, and update-frequency is set to 2000, 10, and 10, according to the recommendations of \cite{slak}. Other settings follow the fine-tuning settings of \cite{slak}. These models follow the training protocol for classification and are not optimized for WSOL (e.g., we do not evaluate WSOL scores at each epoch). And our high-score, the 100-epoch RepLKNet31B1K384 model's 90.99\%MaxBoxAcc in the CUB-200-2011 dataset, actually reports the best score of the checkpoint models saved every 50 epochs.

\begin{figure}[t]
  \centering

    \includegraphics[width=1\linewidth]{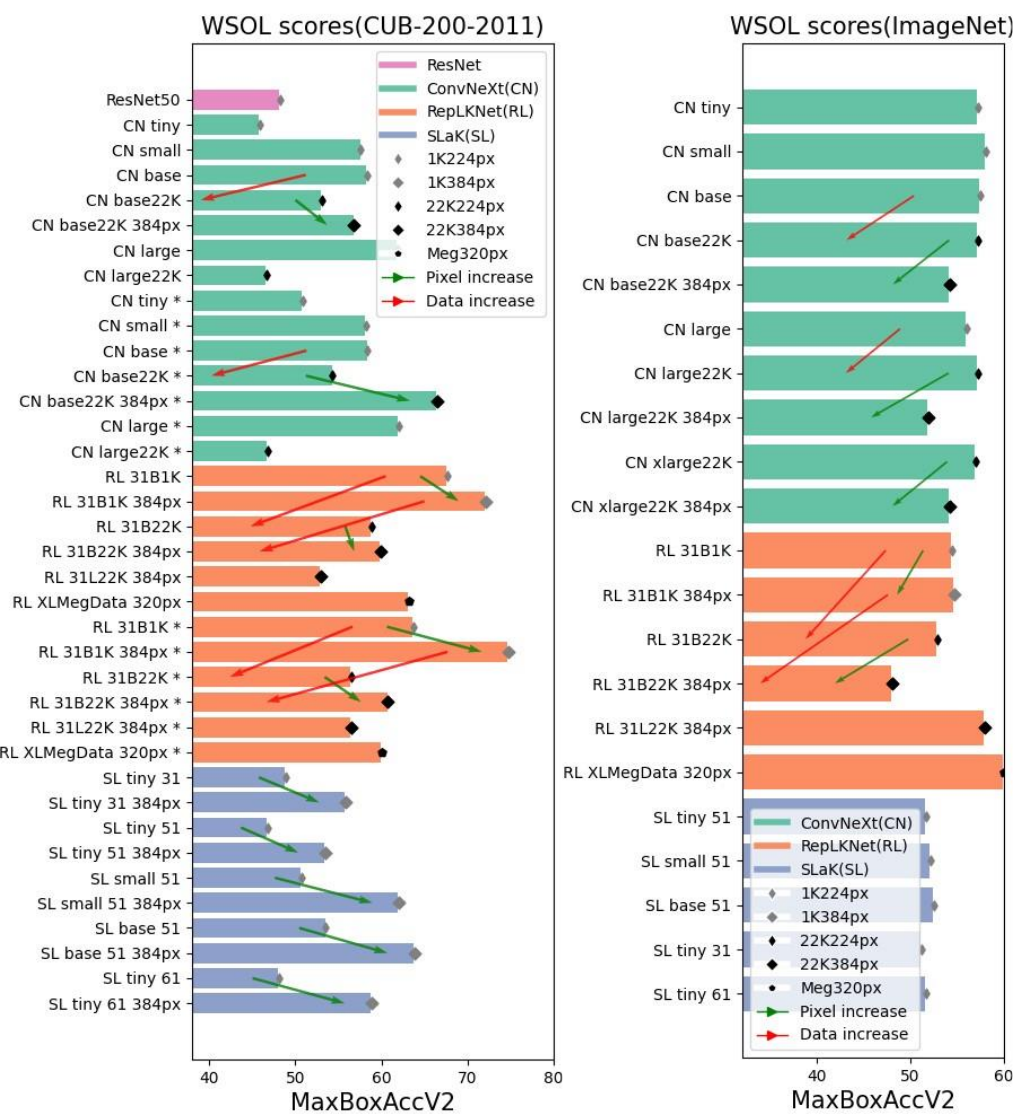}
    
   \caption{
   WSOL scores (MaxBoxAccV2) for the CUB-200-2011 dataset on the left and the ImageNet1K dataset on the right. Green arrows connect models that differ only in training pixel size settings, and red arrows connect models that differ only in pre-training settings. * denotes models that introduced the same data augmentation strategy as the pre-training in fine tuning.
    }
   \label{fig:apfig001}
\end{figure}

\begin{figure}[t]
  \centering

    \includegraphics[width=1\linewidth]{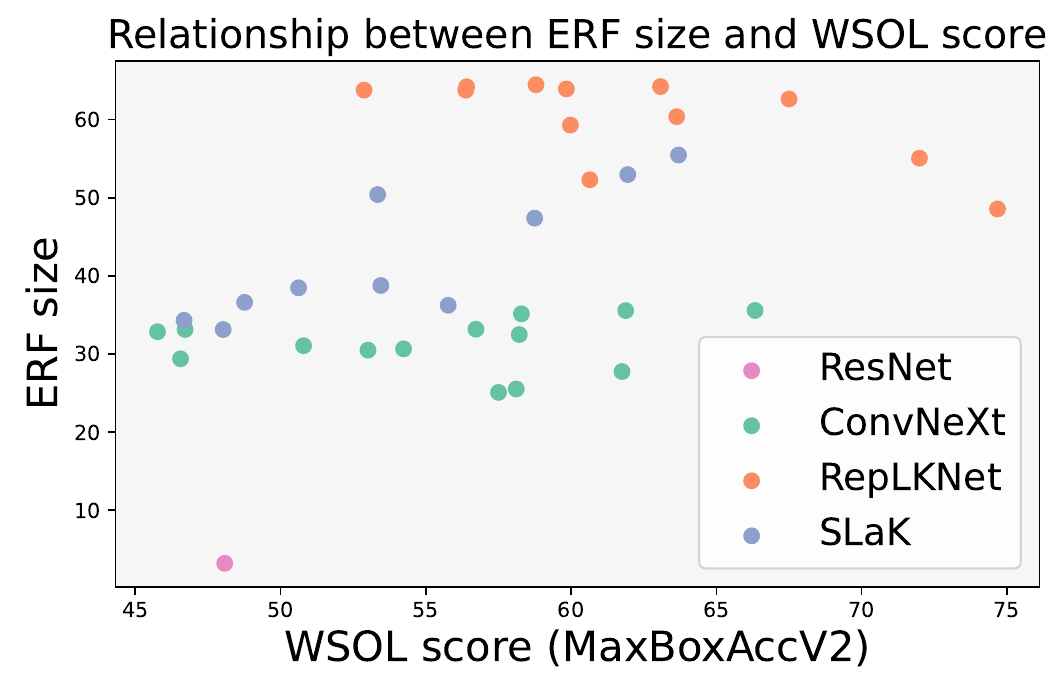}
    
   \caption{
   Graph shows the relationship between WSOL score (MaxBoxAcc2) and ERF size. Here, ERF size is measured by the Area Under the Curve (AUC) of \cref{fig:fig010} (a).
    }
   \label{fig:apfig002}
\end{figure}

\section{WSOL Evaluation Details}
\label{sec:app_wsol_eval}

Our WSOL evaluation methodology is largely based on the \cite{eval_wsol_right}. For the CUB-200-2011 dataset, we use 5994 images (train-weaksup) for training (fine-tuning) and 5774 images (test) for testing, as well as 1000 images (train-fullsup) for threshold search. 
There is a problem with the train-fullsup label information provided in the ImageNet1K (ILSVRC2012) dataset
\cite{git_eval_wsol_right}. Therefore, threshold search and testing will be performed by a test set of 50,000 images. Apart from MaxBoxAcc, which is primarily reported in this paper, here we present the results of MaxBoxAccV2 (\cref{fig:apfig001,fig:apfig002}). Checking the direction of the arrows, we can see the same trend as in the MaxBoxAcc graph \cref{fig:apfig001}.

\begin{figure*}
  \centering

    \includegraphics[width=1\linewidth]{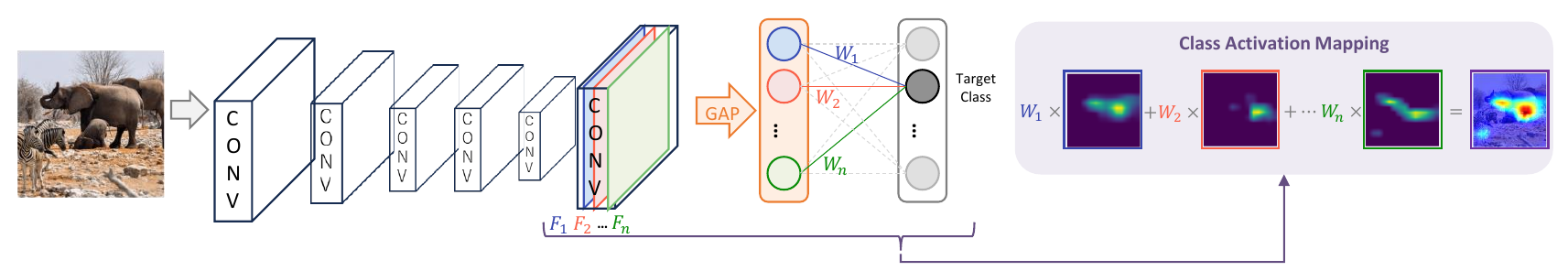}
    
   \caption{
   Calculation process of CAM. To calculate CAM, the feature maps of the final convolution layer and the weights of the FC layer, which connect the feature vectors obtained by global average pooling of those maps to the final logit, are used.
    }
   \label{fig:apfig005}
\end{figure*}

\section{Analysis on ERF and Shape Bias}
\label{sec:app_shape_bias}

\begin{figure}[t]
  \centering

    \includegraphics[width=1\linewidth]{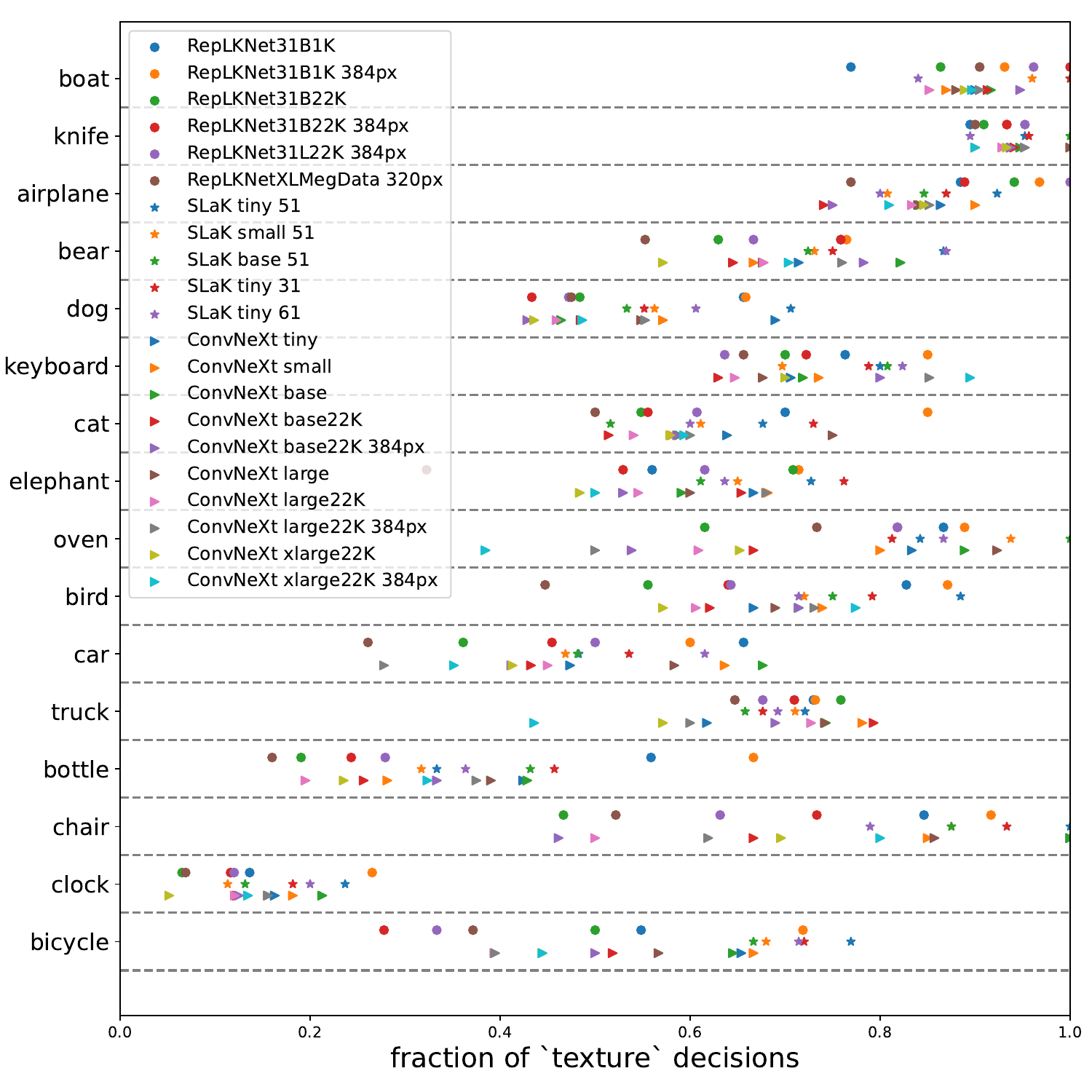}
    
   \caption{
   Texture bias and shape bias measurements in the latest CNNs.
    }
   \label{fig:apfig003}
\end{figure}

The latest research \cite{shape_bias} argues that Vision Transformers are more human-like in that they tend to make predictions based on the overall shape of objects, whereas CNN is more likely to be based on local texture. In contrast, \cite{replknet} show that RepLKNet has a higher shape bias than Swin Transformer and small kernel CNNs. Furthermore, from the perspective of comparison with the Swin Transformer, they argue for the possibility that the strength of the shape bias is closely related to ERFs. Therefore, we performed similar experiments on untested RepLKNet and other latest CNNs by \cite{replknet} using the toolbox \cite{Toolbox} (\cref{fig:apfig003}). \cref{fig:fig010} (a) shows that ERF is larger for ConvNeXt, SLaK, and RepLKNet, but \cref{fig:apfig003} shows that the strength of shape bias is stronger for SLaK, ConvNeXt, and RepLKNet, in that order. These facts negate any simple correlation between the strength of the shape bias and ERF. Also, considering our findings in this paper, the strength of the shape bias seems to be related to the model's capability to generate feature maps with large activation regions.

\begin{figure}[t]
  \centering

    \includegraphics[width=1\linewidth]{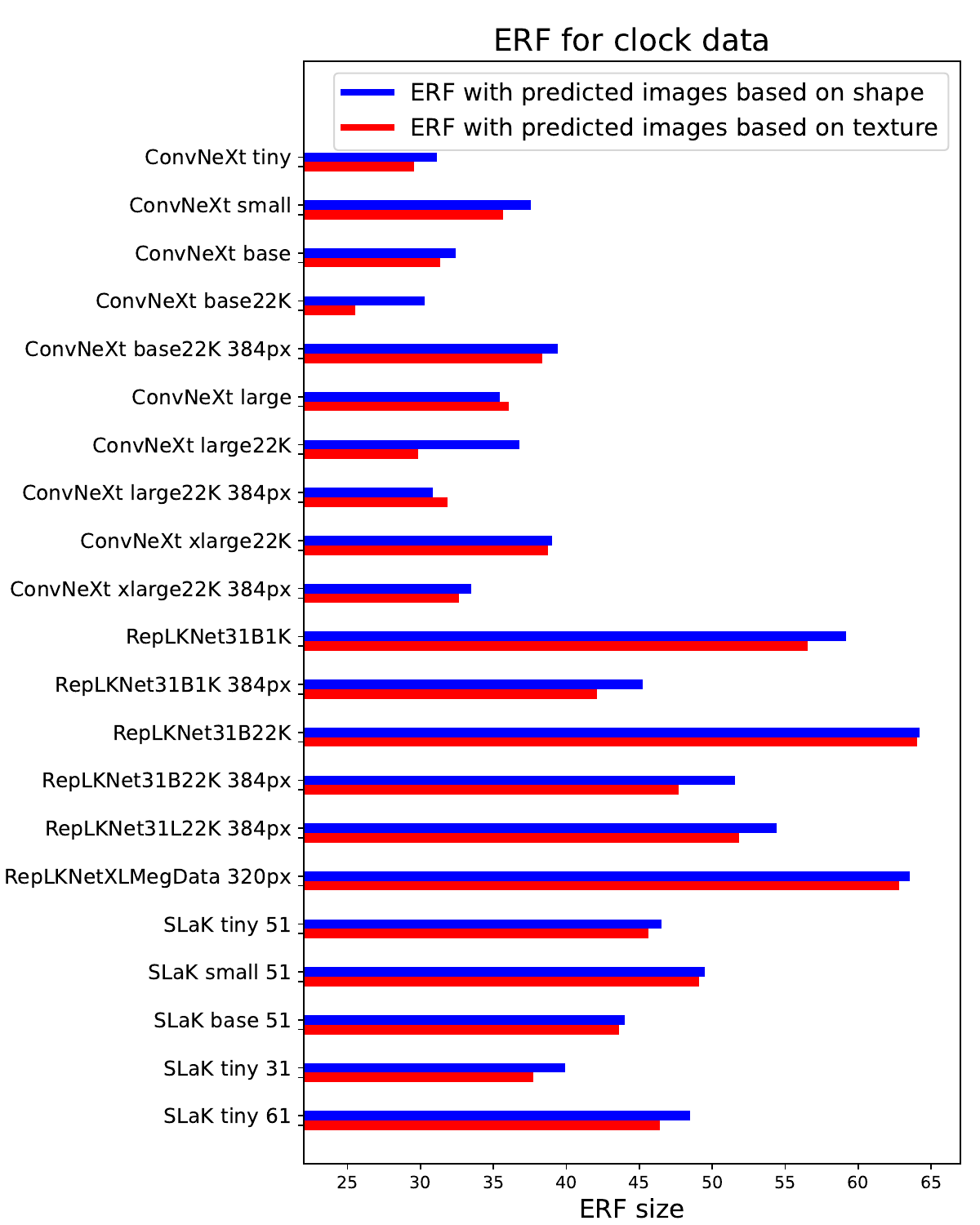}
    
   \caption{
   Graph comparing ERF sizes obtained from different image groups. For images whose shape is clock and whose style is the other category, the ERF sizes obtained from the group of images predicted to be in the clock category are shown in blue. The ERF sizes obtained from the group of images predicted to be in the style side category are shown in red. The calculation of ERF size is the same as in \cref{fig:fig010} (a).
    }
   \label{fig:apfig004}
\end{figure}

On the other hand, from a more microscopic perspective, the relationship between shape bias and ERFs can be seen. As an example, consider the clock category in  \cref{fig:apfig003}. Clock is a category with strong shape bias in many models---an image whose shape is a clock and whose style is another category (e.g., elephant) is likely to be predicted as a clock category. Among the images whose shape is a clock, the ERFs obtained from the group of images predicted to be in the clock category and predicted to be in the style side category show that the former ERF is larger for most models (\cref{fig:apfig004}). 
These results can provide clues to understanding the perceived usefulness of large ERFs, including long-range dependencies.

\section{CAM(Class Activation Map) Calculation Method and Semantic Interpretation}
\label{sec:app_cam_calc}

Class Activation Map (CAM) is a method for localizing object regions in an image that belong to a specific category using CNN with Global Average Pooling (GAP). It is a typical method in WSOL tasks and is often used as a baseline method. To calculate CAM, the feature maps of the final convolutional layer and the weights of the FC layer that connect the feature vectors obtained by GAP of those maps to the final logit are used (\cref{fig:apfig005}). The feature maps of the final convolutional layer is $F$, the $n$-th channel slice is $F_n\in\mathbb{R}^{I\times J}$, and the size of the feature map $F_n$ is $I\times J$. The feature map $F$ in original CNN architecture is transformed into a feature vector $G\in\mathbb{R}^N$ by GAP layer. The scalar $G_n$ corresponding to the feature map $F_n$ is 
the glocal average pooling of $F_n$
\begin{equation}
  G_n = \frac{1}{I\times J}\sum_{i}\sum_{j}F_n (i,j).
  \label{eq:eq001}
\end{equation}

The feature vector $G$ obtained here is transformed into a C-dimensional logit vector through the final FC layer ($C$ is the number of classes). 
Let $W$ be the weight of the FC layer of size $N\times{C}$. The weight between the feature scalar $G_n$ and the final logit for the class $c$ is  $W_{n,c}\in\mathbb{R}$. The class activation map $M_c\in\mathbb{R}^{I\times J}$ for class $c$ is then given by the following weighted sum over the channel dimension
\begin{equation}
  M_c = \sum_{n=1}^{N}W_{n,c}\cdot{F_n}.
  \label{eq:eq002}
\end{equation}

The semantic interpretation of the CAM is as follows. In the classification model, the $N$ weights between the $N$ elements of feature vector $G\in\mathbb{R}^N$ and the final logit for a given class $c$ can be interpreted as carrying information about how important each feature element $G_n\in\mathbb{R}$ should be to obtain correct class predictions. In addition, spatially meaningful information is stored in the feature map of the CNN. Therefore, by assigning this weight as the importance of each feature map $F_n\in\mathbb{R}^{I\times J}$, spatially meaningful regions emerge from the feature map set.

In the WSOL task, binarization by threshold is used to obtain the prediction results of localization from the generated heatmap as a bounding box. According to the \cite{eval_wsol_right}, the determination of the threshold value should be calculated using a different dataset (fullsup dataset) than the data for classification model training or the final prediction. The method of obtaining the final predicted regions from the binarized map depends on the evaluation method (MaxBoxAcc or MaxBoxAccV2).

\section{Details of Problems Faced by CAM in WSOL Tasks}
\label{sec:app_cam_pbm}

This chapter provides a detailed description of the problems CAM faces in the WSOL task shown in \cref{fig:fig002}. The two problems shown in the figure are also discussed in \cite{rethinking_cam}.

First, for the three feature maps and weights indicated by $F_i$ in the upper row in \cref{fig:fig002}. In training for classification, the smaller (larger) the activation area of a feature map, the larger (smaller) the weight tends to be, and $F_i$ in the figure represents such a state. If the CAM is calculated under such conditions, it will look like the CAM in the upper right corner. The central feature map, which has a small activation area but large weights, has a significant contribution to the generated CAM, resulting in a CAM in which the bird's head is particularly activated. For clarity, the CAM is generated here from only three feature maps corresponding to positive weights ($F_{pos}$), but this is a known problem that occurs regardless of the plus/minus of the weights or the number of maps.

Next, for the three feature maps and weights indicated by $F_j$ in the bottom row. These feature maps activate portions of the object region that are not important for classification. The activation of feature maps corresponding to negative weights (hereafter referred to as $F_{neg}$) should ideally not be activated in the classification task, since it only reduces the activation value of the final logit for the correct target class. However, $F_{neg}$ are generally also activated. In this case, the activated region should be at least a no-object region to avoid adversely affecting the shape of the CAM. It is however often activated within an object region, such as $F_j$. $F_{neg}$ works to deactivate those activated regions when computing the CAM. In this example, the areas of the bird's abdomen and feet are deactivated. Thus, when the CAM for $F_i$, which was already locally activated, is combined with the CAM for $F_j$, it becomes an even more locally activated CAM (bottom right CAM).

\section{More Analysis of Activation Area Size and Weight Size}
\label{sec:app_fmap_ana}

\begin{figure}[t]
  \centering

    \includegraphics[width=1\linewidth]{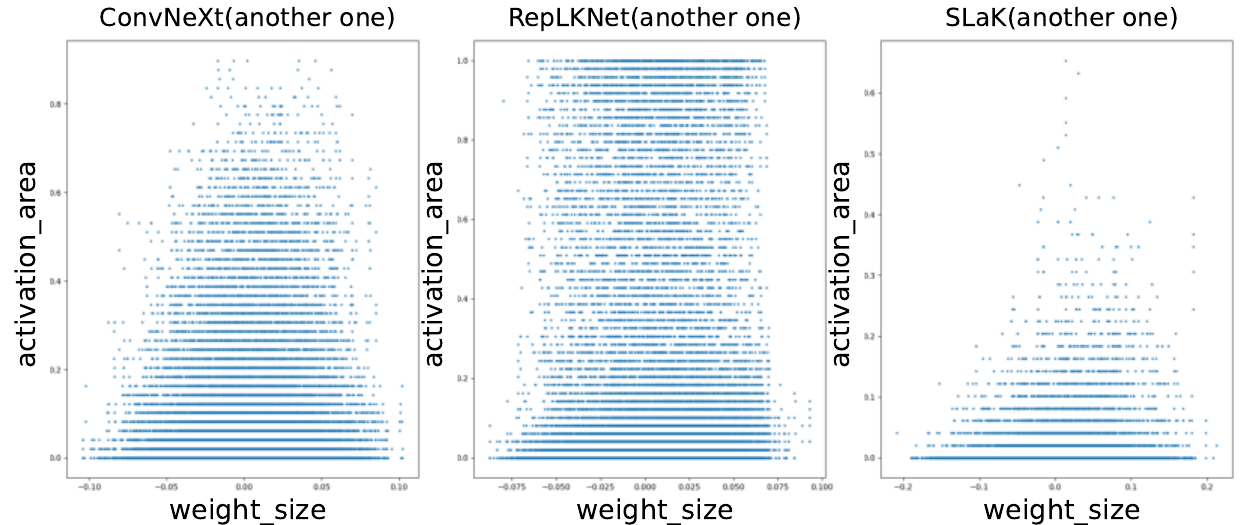}
    
   \caption{
   Relationship between activation areas of the feature maps and weights. The activation area is calculated by binarizing the feature map with a threshold of 10 and calculating the percentage of pixels that exceed the threshold. Experiments on ConvNeXt, RepLKNet, and SLaK, fine-tuned on the CUB-200-2011 dataset.
    }
   \label{fig:apfig006}
\end{figure}

Each plot in \cref{fig:apfig006} shows the relationship between activation areas of the feature maps and weights. This is the same experiment as in \cref{fig:fig005}, but using a model that did not have the best WSOL score. The model identifiers used are \verb^ConvNeXt_tiny^, \verb^RepLKNet31B22K224^, and \verb^SLaK_tiny51_224^. The model identifiers with the best WSOL scores (\cref{fig:fig005}) are \verb^ConvNeXt_base_384_in22ft1k^, \verb^RepLKNet31B1K384^, and \verb^SLaK_base51_384^.

\begin{figure}[t]
  \centering

    \includegraphics[width=1\linewidth]{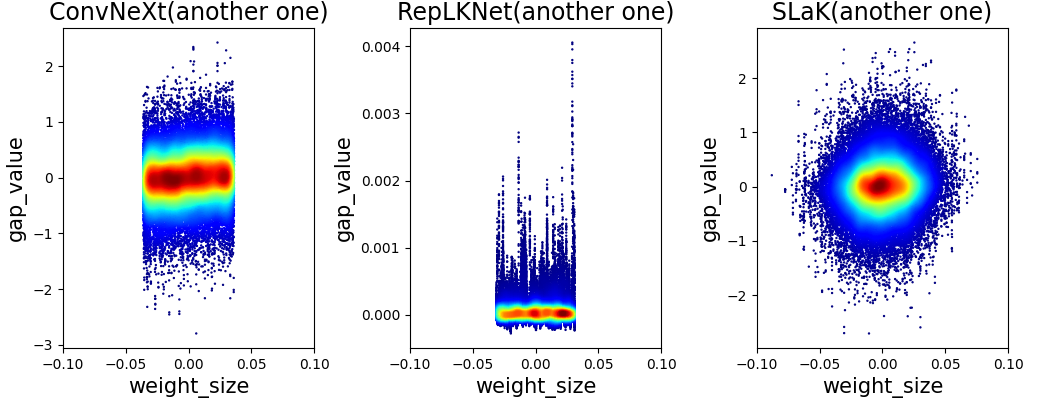}
    
   \caption{
   Relationship between GAP values of the feature maps and weights. Experiments on ConvNeXt, RepLKNet, and SLaK. Models with non-best WSOL scores in CUB-200-2011 dataset are used and the weights are randomly initialized. Plotted against 100 groups of feature maps obtained from 100 random images selected from the same dataset.
    }
   \label{fig:apfig007}
\end{figure}

\cref{fig:apfig007} shows the relationship between GAP values of the feature maps and weights. This is the same experiment as in \cref{fig:fig012}, but using a model that did not have the best WSOL score. The non-best scoring model used are the same as in \cref{fig:apfig006}.

Each figure shows that similar trends discussed in the main text can be observed even when the model is changed.

\section{Quantification of Feature Maps Complexity}
\label{sec:app_fmap_complexity}

\begin{figure}[t]
    \vspace{-1em}
    \centering

    \includegraphics[width=1\linewidth]{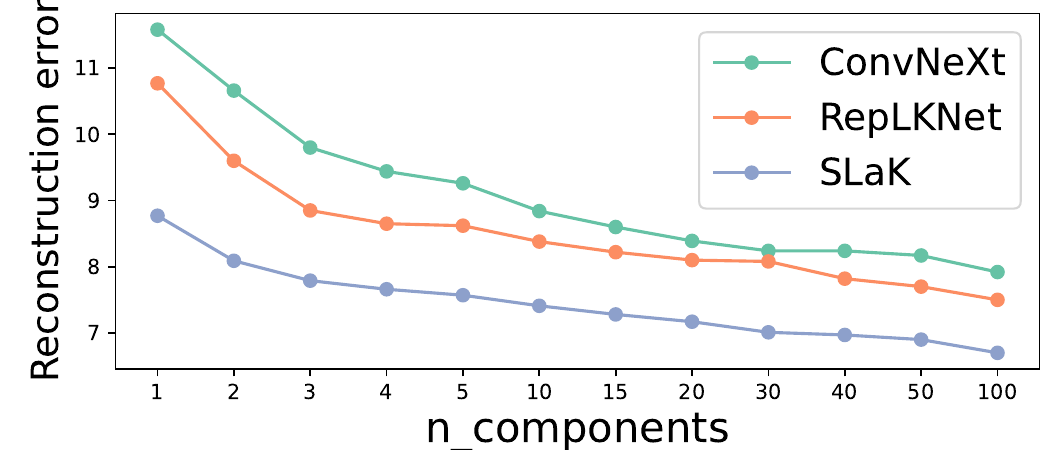}
    
    \vspace{-1em}
    \caption{ 
    Reconstruction error by dictionary learning.
    }
    \label{fig:fig_ext001}
        \vspace{-1.5em}
\end{figure}

For the reasons discussed in \cref{sec:diff_in_cam_quality}, we believe that \cref{fig:fig008} is material to quantify the complexity of the feature maps. In addition, for more direct quantification, we performed an analysis using dictionary learning. Specifically, the dictionary is learned from the sampled feature maps using Orthogonal Matching Pursuit method, and feature maps not used for training are reconstructed using the dictionary. The more complex the feature map set, the larger the reconstruction error is expected to be. \cref{fig:fig_ext001} shows the reconstruction errors when the number of components in the dictionary is varied. Regardless of the number of components, the relationship between the three models is constant, with ConvNeXt having the largest error. In addition, the relationship between the three models and the results in \cref{fig:fig008} is consistent and supports our assertion. Note that SLaK feature maps of low complexity are exceptional in quality and size (\cref{fig:fig003}, \cref{fig:fig007}, \cref{tab:table_ext001}).

\section{What Does the Architecture Need to Achieve the Desired Characteristics?}
What is needed in the architecture to get a feature maps like ConvNeXt or RepLKNet? Based on our results so far, we believe it is important to use a simple, large kernel that does not involve decomposition into rectangular kernels or spercity as used in SLaK. This is an insight gained from the fact that SLaK inherits most of its architecture from ConvNeXt, yet does not have the desired characteristics, and the similarities between ConvNeXt and RepLKNet, which have very different architectures. However, answering this question accurately requires a more comprehensive examination and is an interesting future study.

\section{Additional material on CAM quality differences between RepLKNet and ConvNeXt}
\label{sec:sup_quality_difference}

As mentioned in \cref{sec:resistance_}, PC1 in ConvNeXt and RepLKNet seem to be well suited for localization. Therefore, we actually generated a localization map based on the binarized PC1 heatmap and used it to measure WSOL scores (\cref{tab:table_ext001}). As a result, RepLKNet scored higher than ConvNeXt.
Furthermore, ConvNeXt's IoU threshold \cite{eval_wsol_right} is very high (\cref{tab:table_ext001}). The high IoU threshold implicitly indicates the large activation area. For example, if the activation area tends to be larger than the actual object area, the threshold should be higher to round unwanted activation to zero for localization. In fact, when several WSOL methods are applied to ResNet, the optimal IoU threshold is between 10 and 40 \cite{eval_wsol_right}. Additionally, the average optimal thresholds for the CAMs of ConvNeXt, RepLKNet, and SLaK in this paper were 45, 28.5, and 4.1, respectively. Thus, the threshold for ConvNeXt+PC1 is high and can be interpreted as large PC1 activation. This result is another material that the quality of RepLKNet's PC1 is superior to that of ConvNeXt.

Interestingly, the WSOL score based on PC1 is even higher than the CAM score discussed in this paper (\cref{tab:table001}, \cref{fig:fig001}), with a RepLKNet MaxBoxAcc of 93.6\% (\cref{tab:table_ext001}). To our knowledge, this is 0.43\% higher than the state-of-the-art performance of CNN-based WSOL (\cref{tab:table001}). Simply put, this new WSOL method uses PC1 instead of feature maps in CAM. See \cref{sec:pc1_wsol} for details on the methodology.

\section{Details of WSOL Method Using PC1}
\label{sec:pc1_wsol}

This section provides details on the WSOL methodology reported in \cref{sec:diff_in_cam_quality} and \cref{sec:sup_quality_difference}. In this method, localization maps are generated from PC1 features obtained by PCA. As shown in \cref{fig:fig007}, the visualized PC1 features do not tell whether the object region is white or black. This method contains an ingenious solution to this problem. The procedure is as follows. (1) Perform PCA instead of GAP in CAM to obtain PC1 features. (2) Binarize PC1 using the average value of all pixels in PC1 as the threshold value. (3) Calculate the average value of the outer edge pixels of the binary map (e.g., 44 pixels for a 12x12 map). (4) For example, if the average value (3) is close to 0, the object area shall be indicated by 1. This method has some characteristics that should be noted, including the lack of class discriminability and the fact that it is heavily influenced by the quality of the feature maps. However, high WSOL performance can be expected for models with high-quality feature maps. The code for our experiments is available at: \href{https://github.com/snskysk/CAM-Back-Again}{https://github.com/snskysk/CAM-Back-Again}.

\end{document}